\documentclass[conference]{IEEEtran}
\renewcommand\IEEEkeywordsname{Index Terms}
\usepackage{afterpage}
\usepackage{amssymb}
\usepackage{amsmath,amsfonts}
\usepackage{optidef}
\usepackage{physics}
\DeclareMathOperator*{\minimize}{minimize}
\usepackage{bbding}
\usepackage{bbm}
\usepackage{dsfont}
\usepackage{algorithm}
\usepackage{algorithmic}
\usepackage{booktabs}
\usepackage{multicol}
\usepackage{subfigure}
\usepackage{subcaption}
\usepackage{float}
\usepackage{makecell}
\usepackage{setspace}
\usepackage{stmaryrd}
\usepackage{graphicx}
\usepackage{mwe}
\usepackage{array}
\usepackage{multirow}
\usepackage{multicol}
\usepackage[font=footnotesize]{caption}
\usepackage{textcomp}
\usepackage{stfloats}
\usepackage{url}
\usepackage{lettrine}
\usepackage{verbatim}
\usepackage{graphicx}
\usepackage{cite}
\usepackage{ifpdf}
\usepackage{array}
\usepackage[skins]{tcolorbox}
\usepackage{multirow}
\usepackage{xcolor}
\usepackage{bm}
\usepackage{amsthm}
\usepackage{mathtools, nccmath}
\usepackage{chngcntr}
\usepackage{comment}
\usepackage{lipsum}
\DeclareMathOperator*{\argmax}{arg\,max}

\DeclareMathOperator*{\sort}{sort}
\DeclareMathOperator*{\maximize}{maximize}

\usepackage{textcase}

\usepackage{array} 
\theoremstyle{plain}

\newenvironment{sketchofproof}{%
\proof}{\endproof}
\def\BibTeX{{\rm B\kern-.05em{\sc i\kern-.025em b}\kern-.08em
    T\kern-.1667em\lower.7ex\hbox{E}\kern-.125emX}}
\allowdisplaybreaks
\graphicspath{{Figures/}}
\begin{document}

\title{ZorBA: Zeroth-order Federated Fine-tuning of LLMs with Heterogeneous Block Activation}

\author{
\IEEEauthorblockN{Chuiyang Meng\IEEEauthorrefmark{1}, Ming Tang\IEEEauthorrefmark{2}, and Vincent W.S. Wong\IEEEauthorrefmark{1}}
\IEEEauthorblockA{\IEEEauthorrefmark{1}Department of Electrical and Computer Engineering, The University of British Columbia, Vancouver, Canada}
\IEEEauthorblockA{\IEEEauthorrefmark{2}Department of Computer Science and Engineering, Southern University of Science and Technology, China}
\IEEEauthorblockA{Email: \IEEEauthorrefmark{1}\{chuiyangmeng, vincentw\}@ece.ubc.ca,
\IEEEauthorrefmark{2}tangm3@sustech.edu.cn}
}
\maketitle

\begin{abstract}
Federated fine-tuning of large language models (LLMs) enables collaborative tuning across distributed clients.
However, due to the large size of LLMs, local updates in federated learning (FL) may incur substantial video random-access memory (VRAM) usage.
Moreover, frequent model exchange may lead to significant communication overhead.
To tackle these challenges, in this paper we propose ZorBA, a zeroth-order optimization-based federated fine-tuning framework with heterogeneous block activation.
ZorBA leverages zeroth-order optimization to eliminate the storage of gradients at the clients by forward passes.
ZorBA includes a heterogeneous block activation mechanism in which the central server allocates different subsets of transformer blocks to clients in order to accelerate the convergence rate and reduce the VRAM usage.
Furthermore, ZorBA utilizes shared random seeds and the finite differences of gradients in order to reduce the communication overhead.
We conduct theoretical analysis to characterize the effect of block activation decisions on the convergence rate and VRAM usage.
To jointly enhance the convergence rate and reduce the VRAM usage, we formulate an optimization problem to optimize the block activation decisions. 
We propose an $\epsilon$-constraint lexicographic algorithm to solve this problem.
Experimental results show that ZorBA outperforms three federated fine-tuning baselines in VRAM usage by up to 62.41\% and incurs a low communication overhead.

\end{abstract}


\section{Introduction}
\label{sec:introduction}
Large language models (LLMs) \cite{zhao2023survey, achiam2023gpt, grattafiori2024llama, team2024gemma, 11044734} have shown remarkable performance across numerous natural language processing tasks.
Fine-tuning techniques further enhance LLMs' ability to handle specific downstream applications \cite{ding2023parameter, hu2022lora, liu2022few}.
Meanwhile, federated learning (FL) \cite{mcmahan2017communication, li2020federated, karimireddy2020scaffold, wang2020tackling} has emerged as a promising paradigm for collaborative training of machine learning models.
By allowing multiple resource-constrained clients to locally train the models and aggregate their parameters without sharing raw data, FL facilitates the deployment of large-scale machine learning applications across distributed environments.
By combining these two approaches, federated fine-tuning empowers LLMs to adapt to diverse and decentralized datasets across clients \cite{qin2023federated, 10447454, wang2024flora, bai2024federated, 10855336, 11044514, panchal2024thinking}.

However, most conventional FL approaches use backpropagation (BP) to compute the first-order gradients for model updates. 
This leads to two critical challenges for directly applying FL to fine-tune LLMs. 
First, LLMs typically contain a number of transformer blocks, resulting in model sizes ranging from hundreds of millions to billions of parameters.
Storing the corresponding gradients during fine-tuning requires substantial video random-access memory (VRAM) on graphics processing units (GPUs). 
This requirement may exceed the VRAM capacities of resource-constrained clients, thus hindering the deployment of FL for LLM fine-tuning. 
Second, conventional FL cannot be used when the first-order gradients are unavailable, such as in models with non-differentiable operators or in black-box systems. 

\begin{figure}
    \centering
    \includegraphics[width=0.44\textwidth]{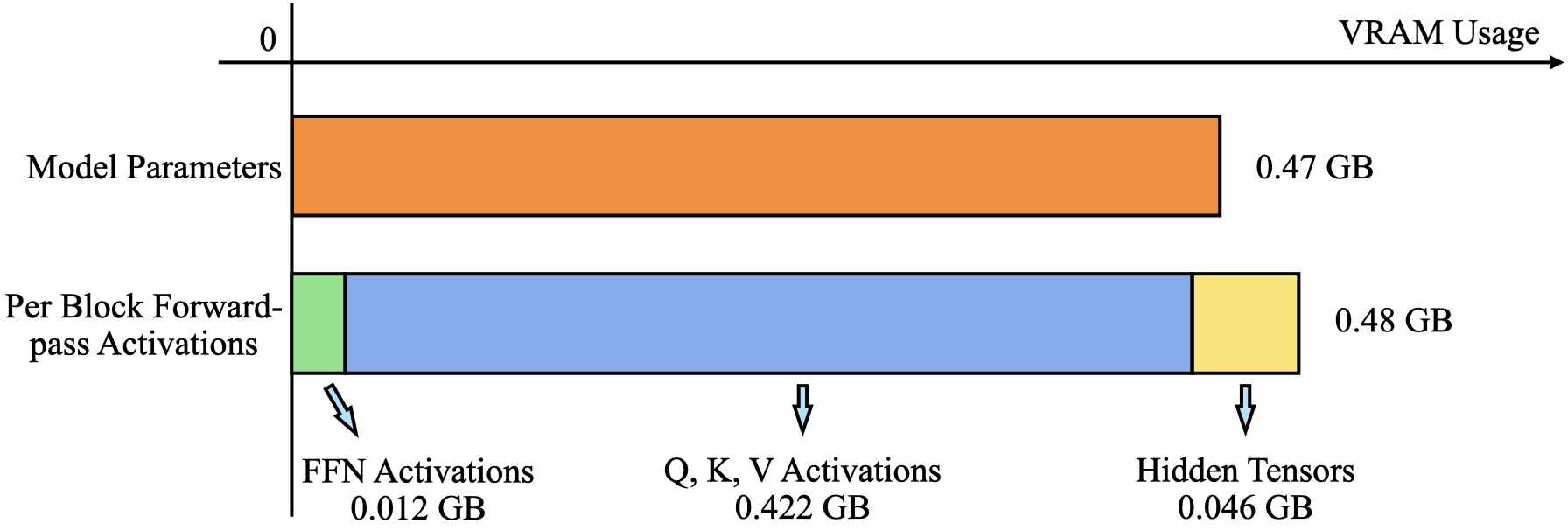}
    \caption{VRAM usage during zeroth-order optimization with OPT-125M.
    The forward-pass activations per block include tensors for hidden tensors, projections for Q, K, and V in transformer blocks, and feed-forward networks (FFNs).}
    \label{fig:vram_usage}
    \vspace{-10pt}
\end{figure}

To address the aforementioned challenges, zeroth-order optimization has emerged as an effective alternative to the conventional first-order FL approaches in LLM fine-tuning \cite{shu2023zeroth, malladi2023fine, chen2023deepzero, zhang2024revisiting, ma2025revisiting, chen2024enhancing}.
In particular, zeroth-order optimization replaces BP-based first-order optimization approaches with a forward-pass-only approach (i.e., a BP-free approach).
Zeroth-order optimization estimates the true gradients by using finite differences of loss function values generated by random perturbation vectors.
Several recent studies have successfully integrated zeroth-order optimization into federated fine-tuning \cite{9917343, chen2023fine, qiu2023zeroth, ling2024convergence, li2024achieving, shu2024ferret}.
Nevertheless, the aforementioned works applied zeroth-order optimization to all blocks, leading to three main limitations.
First, as shown in \cite{malladi2023fine}, zeroth-order optimization approaches exhibit slower convergence rates when compared with first-order approaches in high-dimensional model parameter space.
This is because the perturbation vectors introduce variance on the estimated gradients, which grows with the model dimension.
Second, while zeroth-order approaches eliminate the need to store the gradients at the clients, they still require storing forward-pass activations.
As shown in Fig. \ref{fig:vram_usage}, these activations constitute a significant portion of VRAM usage and increase linearly with the number of activated blocks.
Third, due to frequent client-server communications in FL, the high dimension of the estimated gradients may result in considerable communication overhead.

Based on the aforementioned discussions, we focus on addressing the following question: \textit{Is there a zeroth-order approach enabling clients to activate fewer blocks while reducing the overhead during client-server communications?}

In response, we aim to develop a federated fine-tuning framework integrating zeroth-order optimization with a selective block activation strategy across distributed clients. 
Achieving such a goal is challenging due to the following unexplored questions:
(i) \textit{How can this framework optimally allocate subsets of blocks to clients in order to jointly improve the convergence rate while reducing the VRAM usage across clients?}
(ii) \textit{How can we improve the convergence rate of this federated fine-tuning framework?}
(iii) \textit{Can we propose an efficient and scalable algorithm to determine the block activation decisions under multiple clients and blocks?}
In this work, we make the following contributions as answers to the aforementioned questions:
\begin{itemize}
    \item We propose ZorBA, a zeroth-order federated fine-tuning framework with heterogeneous block activation for LLMs.
    ZorBA reduces the VRAM usage by using zeroth-order optimization.
    It synchronizes the perturbation vectors across the clients and central server via shared random seeds, which prevents gradient leakage and reduces the communication overhead.
    ZorBA includes a heterogeneous block activation mechanism, enabling the clients to activate different subsets of blocks based on the convergence rate and VRAM constraints.

    \item We analyze the convergence rate of ZorBA in the nonconvex setting.
    Through a toy example, we reveal key insights into optimizing block activation decisions.
    We theoretically quantify how these decisions impact the convergence rate and VRAM usage across clients.

    \item We formulate a multi-objective optimization problem to jointly improve the convergence rate and reduce VRAM usage across clients.
    To solve this problem, we propose an $\epsilon$-constraint lexicographic algorithm by decoupling the problem into two subproblems.
    First, we derive a closed-form expression for the maximal least activated block across all clients.
    Then, we design a greedy algorithm under VRAM constraints. 
    It activates additional blocks to minimize the number of clients whose least popularity remains at this maximum.
    We obtain the Pareto front and select the block activation matrix on the front in ZorBA, balancing the convergence rate and total VRAM usage.

    \item We conduct experiments on AG-News, SST-2, and SNLI datasets by using OPT-125M and OPT-1.3B models, respectively.
    We compare ZorBA with FedIT \cite{10447454}, FedZO \cite{9917343}, and DeComFL \cite{li2024achieving}.
    Results show that ZorBA converges faster than FedZO and DeComFL.
    Moreover, ZorBA reduces the communication overhead significantly and reduces the total VRAM usage by up to 62.41\%.

\end{itemize}

The rest of this paper is organized as follows.
Our proposed ZorBA framework is introduced in Section \ref{sec:methodology}.
The convergence rate of ZorBA and theoretical insights are presented in Section \ref{sec:theoretical_analysis}.
The problem formulation and the proposed algorithms are shown in Section \ref{sec:problem_solution}.
Experimental results are presented in Section \ref{sec:performance_evaluation}.
Conclusions are drawn in Section \ref{sec:conclusion}.

\textit{Notations}: We use italic upper case letters, boldface upper case letters, and boldface lower case letters to denote sets, matrices, and vectors, respectively. 
$\mathbb{R}^{M \times N}$ denotes the set of real-valued matrices.
$\mathbf{I}_{d\times d} \in \mathbb{R}^{d\times d}$ denotes the identity matrix.
$\varnothing$ denotes the empty set.
Mathematical operators $\vert\cdot\vert$,
$\mathbb{E}[\cdot]$, $(\cdot)^{\intercal}$, $[\cdot;\cdot]$, $\mathrm{d}(\cdot)$ $\langle\cdot\rangle$, $\lfloor\cdot\rfloor$, , $\Vert\cdot\Vert$, and $\Vert\cdot\Vert_{op}$ denote the cardinality of a set, expectation, transpose, column-wise concatenation, derivative, inner product, floor function, 2-norm, and operator norm of a matrix, respectively.
$\sim$ denotes ``distributed as'' and $\mathcal{U}(\cdot)$ denotes the uniform distribution.

\section{ZorBA}
\label{sec:methodology}
In this section, we present ZorBA, a federated fine-tuning framework.
ZorBA (i) enables heterogeneous block activation in which each client updates different subsets of transformer blocks, (ii) incorporates shared random seeds for independent perturbation vector generation between the clients and central server in order to reduce the communication overhead, and (iii) uses zeroth-order optimization to estimate the gradients with only forward passes in order to reduce the VRAM usage.

\begin{figure}
    \centering
    \includegraphics[width=0.48\textwidth]{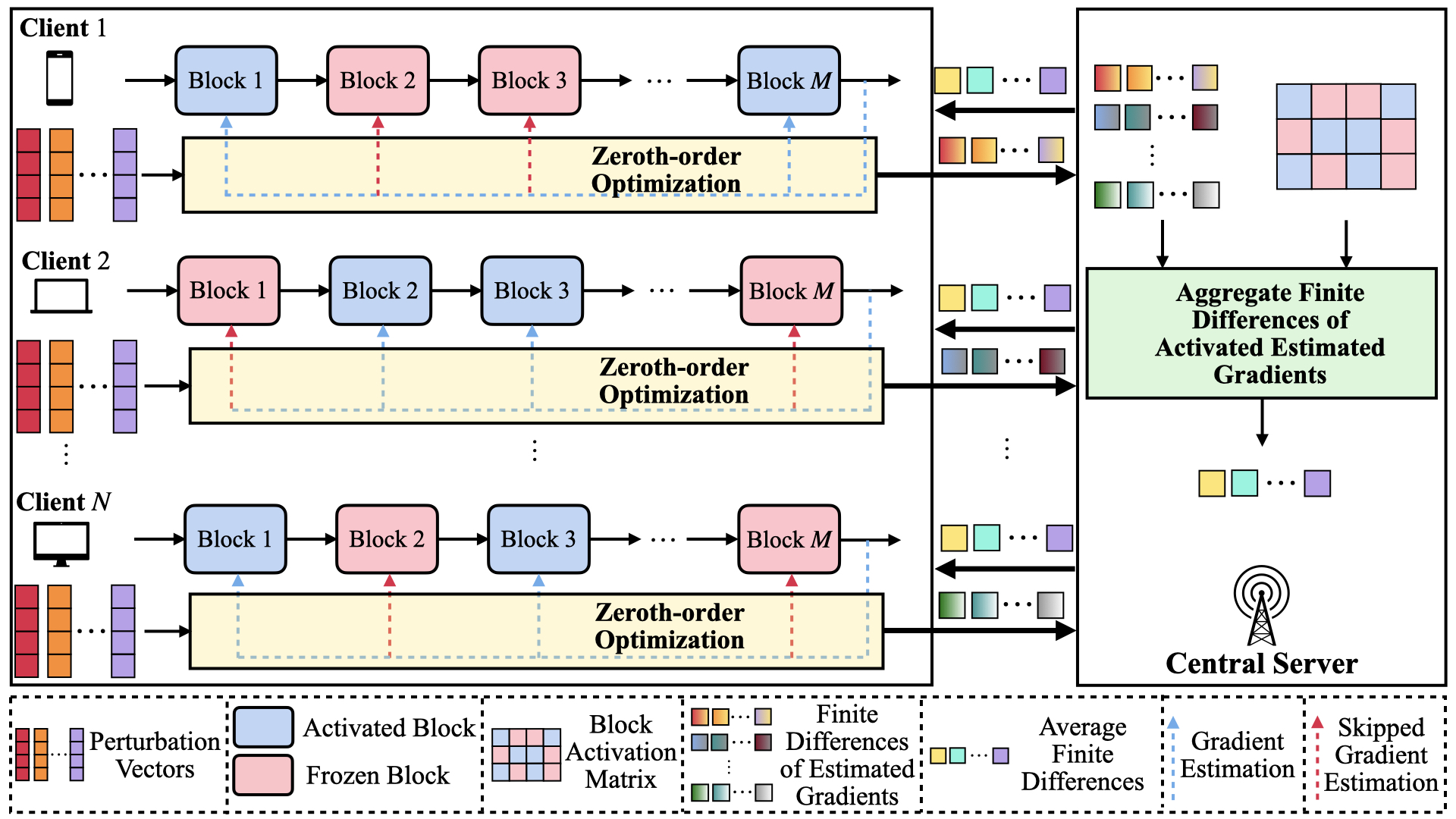}
    \caption{Illustration of our proposed ZorBA framework.}
    \label{fig:system_model}
\end{figure}

We consider a central server and $N$ clients.
Let $\mathcal{T}=\{1,2,\ldots, T\}$ and $\mathcal{N} = \{1, 2, \ldots, N\}$ denote the set of $T$ fine-tuning rounds and the set of $N$ clients, respectively.
Let $\mathcal{M}=\{1,2,\ldots, M\}$ denote the set of $M$ transformer blocks of the pretrained model.
Note that each transformer block is a matrix of model parameters.
For illustration simplicity, we flatten all $M$ blocks of the pretrained model and concatenate them into one vector as $\mathbf{w} = [\mathbf{w}_{1}; \mathbf{w}_{2}; \ldots; \mathbf{w}_{M}] \in\mathbb{R}^{d}$, where $d$ is the model dimension.
Let $\mathbf{w}_{m}$ denote the model parameters of the $m$-th block, $m\in\mathcal{M}$.
We consider that each client $n \in \mathcal{N}$ has a local dataset $\mathcal{D}_{n}$.
Let $\bm{\xi}_{n} \sim \mathcal{D}_{n}$ denote a mini-batch of training samples of $\mathcal{D}_{n}$.
Let $\mathcal{D}$ denote the dataset of all clients.
The central server broadcasts $\Bar{\mathbf{w}}^{1}$ to all clients at the beginning of the first fine-tuning round.
The ZorBA framework is shown in Fig. \ref{fig:system_model}.

\noindent\textbf{Heterogeneous Block Activation:}
In practical systems, clients often have limited VRAM when fine-tuning large models. 
In addition, exchanging the full model update between the clients and central server may result in gradient leakage and privacy concerns.
Therefore, in ZorBA, a heterogeneous block activation mechanism is introduced.
In particular, based on the VRAM capacity of each client, the central server activates a subset of blocks for each client to update.
Let $\mathbf{a}_{n} \in \{0,1\}^{M}$ denote the block activation decision vector for client $n$.
we have $a_{m,n} = 1$ if the $m$-th block is activated for client $n$;
otherwise, $a_{m,n} = 0$.
Let $\mathbf{A} = [\mathbf{a}_{1}, \mathbf{a}_{2}, \cdots, \mathbf{a}_{N}] \in \{0, 1\}^{M\times N}$ denote the block activation allocation decision matrix.
Client $n$ utilizes its decision vector $\mathbf{a}_{n}$ to determine the activated blocks of the local model.
At the beginning of the $t$-th training round, all clients and the central server have the same model, which is obtained at the end of the previous fine-tuning round.
It can be expressed in a ``block form" as $\Bar{\mathbf{w}}^{t} = [\Bar{\mathbf{w}}_{1}^{t}; \Bar{\mathbf{w}}_{2}^{t}; \ldots; \Bar{\mathbf{w}}_{M}^{t}]\in\mathbb{R}^{d}$. 
We denote the parameters of the $m$-th block of client $n$ after block activation as $\Hat{\mathbf{w}}_{m,n}^{t} = a_{m,n}\Bar{\mathbf{w}}_{m}^{t}, m\in\mathcal{M}, n\in\mathcal{N}$.
The activated model parameters of client $n$ in the $t$-th fine-tuning round are written as $\Hat{\mathbf{w}}_{n}^{t} 
    =\: [\Hat{\mathbf{w}}_{1,n}^{t}; \Hat{\mathbf{w}}_{2,n}^{t}; \ldots; \Hat{\mathbf{w}}_{M,n}^{t}]
    =\: [a_{1,n}\Bar{\mathbf{w}}_{1}^{t}; a_{2,n}\Bar{\mathbf{w}}_{2}^{t}; \ldots; a_{M,n}\Bar{\mathbf{w}}_{M}^{t}]\in\mathbb{R}^{d}$.

\noindent\textbf{Shared Random Seeds:}
Due to the large model size, exchanging parameters between clients and the central server incurs significant communication overhead. 
Hence, ZorBA introduces a shared random seed mechanism that replaces model exchange with seed exchange.
At the beginning of fine-tuning, the central server initializes a set of $P$ random seeds $\mathcal{S} = \{s_1, s_2, \ldots, s_P\}$ and shares it with all clients. 
These random seeds are used to initialize random number generators which sample perturbation vectors $\mathcal{V} = \{\mathbf{v}_1, \mathbf{v}_2, \ldots, \mathbf{v}_P\}$ from a standard Gaussian distribution with zero mean and an identity matrix as the covariance matrix \cite{panchal2024thinking}. 
Then, these perturbation vectors are normalized onto the unit sphere. 
Each perturbation vector in set $\mathcal{V}$ will be used for gradient estimation in zeroth-order estimation.
To match the block structure of $\Hat{\mathbf{w}}_{n}^{t}$, each perturbation vector is written as $\mathbf{v}_p = [\mathbf{v}_{p,1}; \mathbf{v}_{p,2}; \ldots; \mathbf{v}_{p,M}] \in \mathbb{R}^d$, where $\mathbf{v}_{p,m}$ is the perturbation vector for the $m$-th block initialized by random seed $s_{p}$.
Since all clients and the central server share the same set of random seeds, they can independently generate the same perturbation vectors based on the exchanged random seeds.

\noindent\textbf{Fine-tuning Process:}
Let $F_{n}(\Hat{\mathbf{w}}_{n}^{t}; \bm{\xi}_{n})$ denote the local loss function of client $n\in\mathcal{N}$ on  $\bm{\xi}_{n}$ with model $\Hat{\mathbf{w}}_{n}^{t}$.
In the $t$-th fine-tuning round, the central server randomly selects a subset of $Q$ seeds from set $\mathcal{S}$ and sends them to all clients. 
Each client then determines the corresponding set of $Q$ perturbation vectors, denoted by $\mathcal{Q}^{t} \subseteq \mathcal{V}$, for zeroth-order optimization.
Client $n\in\mathcal{N}$ determines  $F_{n}(\Hat{\mathbf{w}}_{n}^{t}; \bm{\xi}_{n})$, and then estimates the gradient of $\Hat{\mathbf{w}}_{n}^{t}$ via zeroth-order optimization by $Q$ trials, each corresponding to a perturbation vector.
Note that client $n$ estimates only the gradients of model parameters in the activated blocks.
For the frozen blocks, the corresponding gradient estimation is skipped.
In each trial, client $n$ estimates the gradient by performing two forward passes that evaluate the local loss at the current point and at a perturbed point. 
This is done by adding a sampled perturbation vector $\mathbf{v}_{q} \in \mathcal{Q}^{t}$, scaled by a small smoothing constant $\mu$, to the activated model parameters $\Hat{\mathbf{w}}_{n}^{t}$. 
The constant $\mu$ controls the finite difference of the perturbation applied for gradient estimation.
Then, client $n$ calculates two losses as $F_{n}(\Hat{\mathbf{w}}_{n}^{t} + \mu\mathbf{v}_{q}; \bm{\xi}_{n})$ and $F_{n}(\Hat{\mathbf{w}}_{n}^{t}; \bm{\xi}_{n})$, respectively.
The finite difference of the estimated gradient of client $n$ in the direction of the perturbation vector $\mathbf{v}_{q}$ in the $t$-th fine-tuning round is given by $\rho_{\mathbf{v}_{q}, n}^{t} = \frac{F_{n}(\Hat{\mathbf{w}}_{n}^{t} + \mu\mathbf{v}_{q}; \bm{\xi}_{n}) - F_{n}(\Hat{\mathbf{w}}_{n}^{t}; \bm{\xi}_{n})}{\mu}$.
By averaging all $Q$ trials, the estimated gradient of the $m$-th block of client $n$ in the $t$-th fine-tuning round can be determined by
\begin{align}
    \label{eq:zo_est_seed}
    \Tilde{\nabla}_{\Hat{\mathbf{w}}_{m,n}^{t}} F_{n}(\Hat{\mathbf{w}}_{n}^{t}; \bm{\xi}_{n}) 
    =&\: \frac{1}{Q}\sum_{\mathbf{v}_{q}\in\mathcal{Q}^{t}} \rho_{\mathbf{v}_{q}, n}^{t}\mathbf{v}_{q,m}. 
\end{align}

After finishing local updates, the clients send the model update to the central server for aggregation.
Instead of directly transmitting the full estimated gradient, client $n$ transmits the finite differences of the estimated gradients corresponding to the sampled perturbation vectors.
Since both the model $\Bar{\mathbf{w}}^{t}$ and the sampled perturbation vector subset $\mathcal{Q}^{t}$ are identical between the clients and central server, transmitting only the finite difference of the estimated gradient is sufficient for the server to determine the estimated gradient.
Upon receiving the finite differences of the estimated gradients from all clients, the central server updates the global model based on the estimated gradients over the activated blocks.
In particular, the $m$-th block is updated as follows:
\begin{align}
    \label{eq:model_agg}
    \Bar{\mathbf{w}}_{m}^{t+1} 
    =&\: \sum\limits_{n\in\mathcal{N}}\frac{a_{m,n}}{\sum\limits_{n'\in\mathcal{N}}a_{m,n'}}\left(\Bar{\mathbf{w}}_{m}^{t} -\eta^{t}\Tilde{\nabla}_{\Hat{\mathbf{w}}_{m,n}^{t}} F_{n}(\Hat{\mathbf{w}}_{n}^{t}; \bm{\xi}_{n})\right) \nonumber\\
    =&\: \Bar{\mathbf{w}}_{m}^{t} - \eta^{t}\Big(\frac{1}{Q}\sum\limits_{n\in\mathcal{N}}\frac{a_{m,n}}{\sum\limits_{n'\in\mathcal{N}}a_{m,n'}}\sum_{\mathbf{v}_{q}\in\mathcal{Q}^{t}} \rho_{\mathbf{v}_{q}, n}^{t}\mathbf{v}_{q,m}\Big) 
    \nonumber\\
    =&\: \Bar{\mathbf{w}}_{m}^{t} - \eta^{t}\sum_{\mathbf{v}_{q}\in\mathcal{Q}^{t}}\Bar{\rho}_{\mathbf{v}_{q}}^{t}\mathbf{v}_{q,m}, \quad m\in\mathcal{M},
\end{align}
where $\eta^{t}$ denotes the learning rate in the $t$-th fine-tuning round.
$\Bar{\rho}_{\mathbf{v}_{q}}^{t} = \frac{1}{Q}\sum_{n\in\mathcal{N}}\frac{a_{m,n}}{\sum_{n'\in\mathcal{N}}a_{m,n'}}\rho_{\mathbf{v}_{q}, n}^{t}$ denotes the average finite difference of estimated gradients in the direction of $\mathbf{v}_{q}$. 
Let $\bar{\mathcal{P}}^{t} = \{\Bar{\rho}_{\mathbf{v}_{q}}^{t}: \mathbf{v}_{q}\in\mathcal{Q}^{t}\} $ denote the set of average finite differences for all perturbation vectors in the $t$-th fine-tuning round.
The aggregated model is updated by $\Bar{\mathbf{w}}^{t+1} = [\Bar{\mathbf{w}}_{1}^{t+1}; \Bar{\mathbf{w}}_{2}^{t+1}; \ldots; \Bar{\mathbf{w}}_{M}^{t+1}]$.

Then, the central server broadcasts the set of average finite differences of estimated gradients $\bar{\mathcal{P}}^{t}$ to all clients. 
By using the shared random seeds, the server no longer broadcasts high-dimensional parameters and reduces the communication overhead significantly.
Since both the model $\Bar{\mathbf{w}}^{t}$ and the sampled perturbation vector subset $\mathcal{Q}^{t}$ are shared,
upon receiving $\bar{\mathcal{P}}^{t}$,
each client can obtain the model for the $(t+1)$-th fine-tuning round as $\Bar{\mathbf{w}}^{t+1}$ by using eqn. (\ref{eq:model_agg}).

\noindent\textbf{VRAM Usage Model:}
In zeroth-order optimization, the VRAM usage consists of the storage for both model parameters and forward-pass activations of all blocks. 
We characterize the total VRAM usage as the sum of these two components.
We define $\psi_{\mathrm{max},n}$ as the VRAM capacity for client $n\in\mathcal{N}$.

The VRAM usage for model parameters is identical across clients, as they share the same pre-trained model architecture.
Thus, for any client $n\in\mathcal{N}$ and fine-tuning round $t\in\mathcal{T}$, the VRAM usage for model parameters is denoted as $\psi_{\mathrm{md}}$.

Let $B$, $L$, $H$, and $K$ denote the batch size, input size, hidden size, and number of attention heads, respectively. 
The VRAM usage for forward-pass activations of each block consists of three parts:
(i) hidden state activations:
$\psi_{\mathrm{hd}} = BLH$; 
(ii) activations of Q, K, and V modules in the transformer block: $\psi_{\mathrm{qkv}} = 3KBLH$;
(iii) activations of feed-forward network (FFN): $\psi_{\mathrm{ffn}} = \alpha BLH$, where $\alpha$ is the FFN expansion ratio for the model.
The VRAM usage for each block is given by
$\psi_{\mathrm{block}} = \psi_{\mathrm{hd}} + \psi_{\mathrm{qkv}} + \psi_{\mathrm{ffn}} = (\alpha + 3K + 1)BLH$.
$\psi_{\mathrm{block}}$ models the VRAM usage to store the intermediate states of the blocks at client 
$n$. 
We adopt an additive form since zeroth-order training evaluates multiple perturbations per round and reuses these stored states across evaluations.
For client $n\in\mathcal{N}$ in each fine-tuning round, the VRAM usage of the total activations is the VRAM required to store the activations of all activated blocks, which is given by $\psi_{\mathrm{act}, n}(\mathbf{a}_{n}) = (\alpha + 3K + 1) BLH\sum_{m\in\mathcal{M}}a_{m,n}$.
Thus, the total VRAM usage of client $n\in\mathcal{N}$ is $\psi_{\mathrm{tot}, n}(\mathbf{a}_{n}) = \psi_{\mathrm{act}, n}(\mathbf{a}_{n}) + \psi_{\mathrm{md}}$.

\section{Theoretical Analysis of ZorBA}
\label{sec:theoretical_analysis}
In this section, we analyze how the block activation matrix affects the convergence rate of our proposed ZorBA.
Without loss of generality, we conduct analysis under nonconvex loss functions.
We denote the expected loss of client $n$ with model $\mathbf{w}_{n}^{t}$ as $f_{n}(\mathbf{w}_{n}^{t}) = \mathbb{E}_{\bm{\xi}_{n} \sim \mathcal{D}_{n}}F_{n}(\mathbf{w}_{n}^{t}; \bm{\xi}_{n})$.
We denote the global loss function as $f(\mathbf{w}^{t})$.
We denote the gradient of $f(\mathbf{w}^{t})$ as $\nabla f(\mathbf{w}^{t}) = \mathbb{E}_{n\in\mathcal{N}}\nabla f_{n}(\mathbf{w}^{t})$.
We first present the following assumptions which are widely used in the literature (e.g., \cite{li2019convergence, wang2020tackling, 10971879}).

\noindent\textbf{Assumption 1.} 
\textit{The loss function of each client $n \in \mathcal{N}$ is continuously differentiable and $L$-smooth.
That is, for arbitrary two vectors $\mathbf{w}_{n}^{t}$ and $\tilde{\mathbf{w}}_{n}^{t}$, we have}
$f_{n}(\mathbf{w}_{n}^{t}) \leq f_{n}(\tilde{\mathbf{w}}_{n}^{t}) + \langle \nabla f_{n}(\tilde{\mathbf{w}}_{n}^{t}), \mathbf{w}_{n}^{t} - \tilde{\mathbf{w}}_{n}^{t}\rangle + \frac{L}{2}\Vert\mathbf{w}_{n}^{t} - \tilde{\mathbf{w}}_{n}^{t}\Vert^{2}.$

\noindent\textbf{Assumption 2.} 
\textit{The variance of the local stochastic gradient of each client $n \in \mathcal{N}$ is upper-bounded, i.e., } $\mathbb{E}_{\bm{\xi}_{n}\sim\mathcal{D}_{n}} \big[\Vert\nabla F_{n}(\mathbf{w}_{n}^{t}; \bm{\xi}_{n}) - \nabla f_{n}(\mathbf{w}_{n}^{t}) \Vert^{2}\big] \leq \sigma^{2}$.

\noindent\textbf{Assumption 3.} 
\textit{The dissimilarity between the local gradient of each client $n \in \mathcal{N}$ and the gradient of the global loss function is upper-bounded, i.e., } $\Vert\nabla f(\mathbf{w}^{t}) - \nabla f_{n}(\mathbf{w}^{t}) \Vert^{2} \leq \sigma_{G}^{2}$.

Moreover, we introduce two lemmas to facilitate our convergence analysis.

\noindent\textbf{Lemma 1.}
\textit{Let $\mathbb{E}_{\bm{\xi}_{n}\sim\mathcal{D}_{n}}[\Tilde{\nabla}F_{n}(\mathbf{w}^{t};\bm{\xi}_{n})] = \Tilde{\nabla}f_{n}(\mathbf{w}^{t}), n\in\mathcal{N}$.
The dissimilarity between the zeroth-order and first-order gradient estimation of each client $n \in \mathcal{N}$ is upper-bounded, i.e., }
$\Vert\Tilde{\nabla}f_{n}(\mathbf{w}^{t}) - \nabla f_{n}(\mathbf{w}^{t})\Vert \leq \frac{1}{2}\mu L(d+3)^{\frac{3}{2}}$.
\textit{This result follows from the analysis in} \cite{nesterov2017random}.

\noindent\textbf{Lemma 2.} 
\textit{When $\mu$ is small enough (i.e., $\mu \rightarrow 0$), the square norm of the zeroth-order gradient estimation of each client $n \in \mathcal{N}$ satisfies}
$\mathbb{E}_{\bm{\xi}_{n}\sim\mathcal{D}_{n}}\big[\Vert\Tilde{\nabla} F_{n}(\mathbf{w}_{n}^{t};\bm{\xi}_{n})\Vert^{2}\big] = \frac{d+Q-1}{Q} \mathbb{E}_{\bm{\xi}_{n}\sim\mathcal{D}_{n}}\big[\Vert\nabla F_{n}(\mathbf{w}_{n}^{t};\bm{\xi}_{n})\Vert^{2}\big]$.
\textit{This results from} \cite{malladi2023fine}.

We denote $\Lambda(\mathbf{A}) = \sum_{n\in\mathcal{N}}\max_{m\in\mathcal{M}}\Big\{\Big(\frac{a_{m,n}}{\sum_{n'\in\mathcal{N}}a_{m,n'}}\Big)^{2}\Big\}$.
Let $\mathbf{w}^{\star}$ denote the optimal model.
We present the convergence bound of ZorBA.

\noindent\textbf{Theorem 1.} 
(Standard convergence bound of ZorBA)
\textit{ 
Under Assumptions 1$-$3 and Lemmas 1$-$2, if the learning rate $\eta^{t}$ is small such that $\eta^{t} = \eta \leq \frac{1}{2LdN^{2}}$, the standard convergence bound of our proposed ZorBA satisfies:}
\begin{align}
    \label{eq:converge_std}
    \frac{1}{T}\sum_{t=1}^{T}\left\Vert \nabla f(\Bar{\mathbf{w}}^{t})\right\Vert^{2} 
    \leq&\:
    \frac{1}{\Omega(\Lambda(\mathbf{A}))T}\sum\limits_{t=1}^{T}\left(f(\Bar{\mathbf{w}}^{1}) - f(\mathbf{w}^{\star})\right) \nonumber\\
    &+ \frac{\Theta_{1}}{\Omega(\Lambda(\mathbf{A}))}\Lambda(\mathbf{A}),
\end{align}
where $\Omega(\Lambda(\mathbf{A})) = \left(\frac{\eta}{2} - \frac{3L\eta^{2}(d+Q-1)N}{2Q}\Lambda(\mathbf{A})\right)$ and $\Theta_{1} = \frac{1}{4}\mu^{2}L^{2}\eta(d+3)^{3}N + \left(\eta N + \frac{3L\eta^{2}(d+Q-1)N}{2Q}\right)\sigma_{G}^{2} + \frac{3L\eta^{2}(d+Q-1)N}{2Q}\sigma^{2}$.   

\begin{sketchofproof}
The proof starts by showing the update of one fine-tuning round.
By using Assumption 1, we expand the expected loss of the average model into two terms: a first-order descent term $T_{1}$ and a second-order variance term $T_{2}$.
$T_{1}$ captures the inner product between the true gradient and the model update made with zeroth-order optimization.
In particular, $T_{1}$ is upper-bounded by using the bounded stochastic-gradient variance (Assumption 2), the gradient-dissimilarity assumption (Assumption 3), and the zeroth-order estimation dissimilarity in Lemma 1, giving a bias that depends on the block-activation decisions $\mathbf{A}$.
$T_{2}$ is bounded by using Lemma 2 and Assumptions 2$-$3. 
Lemma 2 draws a connection between the zeroth-order gradients to first-order ones. 
Assumptions 2$-$3 characterize the gradient variance and data heterogeneity, respectively.
By summing over all fine-tuning rounds, rearranging the inequality, and choosing the proper learning rate, the convergence bound is obtained.
\end{sketchofproof}

According to Theorem 1, the convergence bound of ZorBA consists of two terms: an optimality gap that diminishes as the number rounds $T$ increases, and a non-diminishing bias term (i.e., $ \frac{\Theta_{1}}{\Omega(\Lambda(\mathbf{A}))}\Lambda(\mathbf{A})$). 
The bias term quantifies the impact of zeroth-order gradient estimation error, data heterogeneity, and local gradient variance on the convergence. 

However, to guarantee convergence of ZorBA, Theorem 1 requires a learning rate $\eta = \mathcal{O}(\frac{1}{dN^{2}})$. 
In practical systems, the model dimension $d$ can exceed billions, and the number of clients $N$ may be large.
Thus, the required learning rate may be impractically small.
To overcome this challenge, we introduce a condition and adopt a supporting lemma from \cite{malladi2023fine}, allowing us to derive a convergence bound independent of $d$.

\noindent\textbf{Condition 1.} 
\textit{Let $G(\mathbf{w}^{t}) = \max\limits_{n\in\mathcal{N}}\max\limits_{\bm{\xi}_{n}\sim\mathcal{D}_{n}}\Vert\nabla F_{n}(\mathbf{w}_{n}^{t};\bm{\xi}_{n})\Vert$, there exists a Hessian matrix $\mathbf{H}(\mathbf{w}^{t}) \preceq L\mathbf{I}_{d\times d}$ such that for arbitrary $\mathbf{w} \in \mathbb{R}^{d}$ with $\Vert\mathbf{w} -\mathbf{w}^{t}\Vert \leq \eta^{t}dG(\mathbf{w}^{t})$, we have $\nabla^{2}f(\mathbf{w}) \preceq \mathbf{H}(\mathbf{w}^{t})$. 
The effective rank of $\mathbf{H}(\mathbf{w}^{t})$ satisfies $\mathbf{erank}(\mathbf{H}(\mathbf{w}^{t})) = \frac{\Tr(\mathbf{H}(\mathbf{w}^{t}))}{\Vert\mathbf{H}(\mathbf{w}^{t})\Vert_{op}}\leq\kappa$.
This is from} \cite{malladi2023fine}.

Condition 1 shows that the Hessian of local loss can be approximated by a matrix such as its curvature is dominated by only $\kappa$ effective directions.
Hence, the convergence bound depends on $\kappa$ instead of $d$.

\noindent\textbf{Lemma 3.}
\textit{Let $F(\mathbf{w}; \bm{\xi})$ denote the global stochastic loss such that $f(\mathbf{w}) = \mathbb{E}_{\bm{\xi}\sim\mathcal{D}}[F(\mathbf{w};\bm{\xi})]$.
We define $\mathbf{\Sigma}(\mathbf{w}) = \mathbb{E}\left[\nabla F(\mathbf{w}; \bm{\xi})\left(\nabla F(\mathbf{w}; \bm{\xi})\right)^{\intercal}\right] - \nabla f(\mathbf{w})\left(\nabla f(\mathbf{w})\right)^{\intercal}$.
The outer product of the global model difference between two consecutive rounds satisfies}
\begin{align}
    &\mathbb{E}\left[(\Bar{\mathbf{w}}^{t+1} - \Bar{\mathbf{w}}^{t})(\Bar{\mathbf{w}}^{t+1} - \Bar{\mathbf{w}}^{t})^{\intercal}\right] \nonumber\\
    =&\: \Big(1 + \frac{d-2}{d+2}\Big)(\eta^{t})^{2}N\Big(\nabla f(\Bar{\mathbf{w}}^{t})\left(\nabla f(\Bar{\mathbf{w}}^{t})\right)^{\intercal} + \frac{1}{N}\mathbf{\Sigma}(\Bar{\mathbf{w}}^{t})\Big) \nonumber\\
    &+ \frac{d}{d+2}(\eta^{t})^{2}N\Big( \left\Vert\nabla f(\Bar{\mathbf{w}}^{t})\right\Vert^{2} + \frac{1}{N}\Tr\left(\mathbf{\Sigma}(\Bar{\mathbf{w}}^{t})\right)\Big)\mathbf{I}_{d\times d}. 
\end{align}
\textit{This result follows from} \cite{malladi2023fine}.

Then, we present the dimension-free convergence bound of our proposed ZorBA.

\noindent\textbf{Theorem 2.} 
(Dimension-free convergence bound of ZorBA)
\textit{
Under Assumptions 1$-$3, Lemma 1, Condition 1, and Lemma 3, if the learning rate $\eta^{t}$ satisfies $\eta^{t} = \eta \leq \frac{d+2}{2LdN^{2}(\kappa+2)}$, the dimension-free convergence bound of our proposed ZorBA is:}
\begin{align}
    \label{eq:converge_dimfree}
    \frac{1}{T}\sum_{t=1}^{T}\left\Vert\nabla f\left(\Bar{\mathbf{w}}^{t}\right)\right\Vert^{2}
    \leq&\: \frac{1}{\Phi(\Lambda(\mathbf{A}))T}\left(f(\Bar{\mathbf{w}}^{1}) - f(\Bar{\mathbf{w}}^{\star})\right) \nonumber\\
    &+ \underbrace{\frac{\Theta_{2}}{\Phi(\Lambda(\mathbf{A}))}\Lambda(\mathbf{A})}_\text{bias term},
\end{align}
where $\Phi(\Lambda(\mathbf{A})) = \Big(\frac{\eta}{2} - \frac{d\eta^{2}NL(\kappa+2)}{2(d+2)}\Lambda(\mathbf{A})\Big)$ and $\Theta_{2} = \eta N\left(\sigma_{G}^{2} + \frac{1}{4}\mu^{2}L^{2}(d+3)^{3}\right) + \frac{d\eta^{2}L(\kappa+2)\sigma^{2}}{2(d+2)}$.   

\begin{sketchofproof}
To remove the dependency on the model dimension $d$, we expand the expected loss of the average model to second order and invoke the local $\kappa$-effective-rank condition on the Hessian (Condition 1). 
Lemma 3 shows the outer product of model differences, which further connects the Hessian term to the gradient norm and a trace term without $d$.
The key insight is bounding this Hessian term by leveraging the gradient covariance matrix. 
The resulting convergence bound has a similar structure to Theorem 1, but with a new bias term which incorporates $\kappa$ and eliminates the learning rate’s dependence on $d$.

\end{sketchofproof}

Theorem 2 introduces a non-diminishing bias term which reflects the impact of zeroth-order gradient error, data heterogeneity, and local gradient variance while ensuring the learning rate is independent of model dimension. 

To improve convergence, it is essential to design the block activation matrix $\mathbf{A}$ to minimize this bias. 
Although the involved constants (e.g., $L$, $\kappa$, $\sigma_G$, $\sigma$) are generally intractable, Lemma 4 shows that accelerating the convergence is equivalent to minimizing $\Lambda(\mathbf{A})$, regardless of these constants.

\begin{figure}[t]
    \centering
    \includegraphics[width=0.48\textwidth]{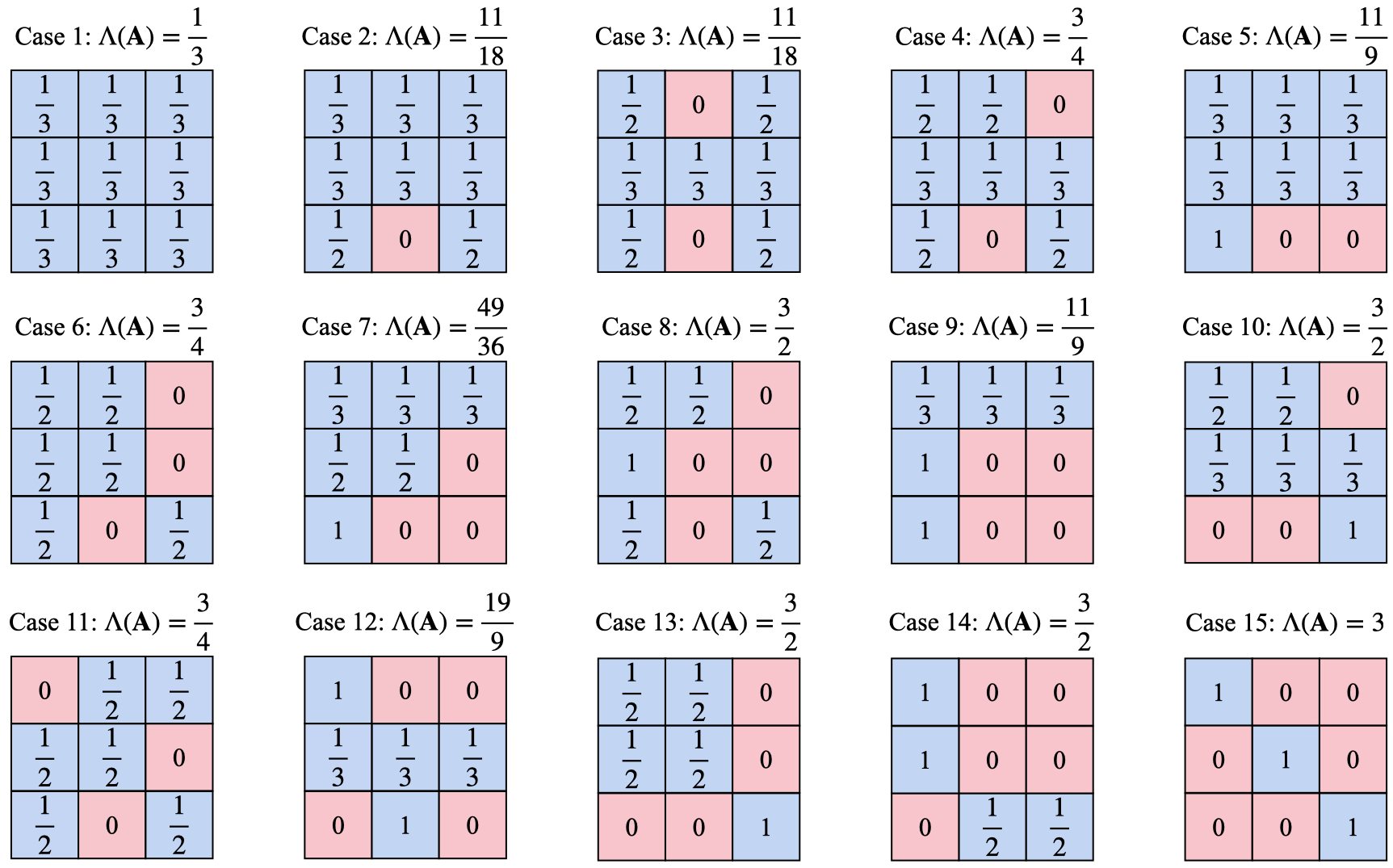}
    \caption{An example of different block activation decisions on three clients. 
    We consider that each client's model has three blocks.
    Each column denotes the blocks for each client.
    The blue square denotes the activated block.
    The pink square denotes the frozen block.
    The value of the block in the $m$-th row and $n$-th column corresponds to the aggregation weight, i.e., $\frac{a_{m,n}}{\sum_{n'\in\mathcal{N}}a_{m,n'}}$.}
    \label{fig:toy_example}
\end{figure}

\noindent\textbf{Lemma 4.} (Monotonicity of bias term)
\textit{The optimal block activation matrix $\mathbf{A}$ which minimizes the bias term $\frac{\Theta_{2}}{\Phi(\Lambda(\mathbf{A}))}\Lambda(\mathbf{A})$ is equivalent to the minimizer of $\Lambda(\mathbf{A})$.}
\begin{proof}
    The first-order derivative of the bias term $\frac{\Theta_{2}}{\Phi(\Lambda(\mathbf{A}))}\Lambda(\mathbf{A})$ with respect to (\textit{w.r.t.}) $\Lambda(\mathbf{A})$ can be derived as 
    \begin{align}
        \frac{\mathrm{d}\frac{\Theta_{2}}{\Phi(\Lambda(\mathbf{A}))}\Lambda(\mathbf{A})}{\mathrm{d}\Lambda(\mathbf{A})} = \frac{\Theta_{2}\Phi(\Lambda(\mathbf{A})) + \frac{d\eta^{2}NL(\kappa+2)}{2(d+2)} \Theta_{2}\Lambda(\mathbf{A})}{\left(\Phi(\Lambda(\mathbf{A}))\right)^{2}}.        
    \end{align}
    In particular, $\Theta_{2}$ and $\Lambda(\mathbf{A})$ are both positive.
    When $\eta \leq \frac{d+2}{2LdN^{2}(\kappa+2)}$, $\Phi(\Lambda(\mathbf{A}))$ is non-negative.
    The bias term is monotonically increasing \textit{w.r.t.} $\Lambda(\mathbf{A})$ and is minimized at the lower bound of $\Lambda(\mathbf{A})$.
\end{proof}

To further demonstrate the impact of the block activation matrix $\mathbf{A}$ on the convergence rate, we present an example in Fig. \ref{fig:toy_example}. 
Based on Lemma 4, we use $\Lambda(\mathbf{A})$ to characterize the bias term and the convergence rate.
We present several important observations as follows.

\noindent\textbf{Observation 1.}
\textit{At a large scale, increasing the number of activated blocks generally reduces $\Lambda$, thereby leading to faster convergence.
The optimal convergence rate is achieved when all clients activate all the blocks, i.e., full-block fine-tuning.}

\noindent\textbf{Observation 2.}
\textit{At a small scale, activating more blocks does not always improve the convergence.}

\noindent\textbf{Observation 3.}
\textit{In some cases, activating more blocks can degrade the convergence.
For example, case 5 activates one more block than case 11, but case 11 has a lower value of $\Lambda$.}

\noindent\textbf{Observation 4.}
\textit{The convergence rate can vary even when the total number of activated blocks remains the same.
For example, both cases 8 and 9 activate the same total number of blocks. 
Although case 9 has a more imbalanced block allocation, it achieves a lower value of $\Lambda$. }



Based on the above observations, we provide insights into minimizing $\Lambda(\mathbf{A})$.
We define $c_{m}(\mathbf{a}_{m,:}) = \sum_{n\in\mathcal{N}}a_{m,n}$ as the popularity of block $m$.
It is the number of clients that activate block $m$.
For each client $n$, we define its least popularity as the minimum popularity among all blocks activated by client $n$.
The least popularity of client $n$ satisfies $\underline{c}_{n}(\mathbf{A}) = \min\limits_{m:a_{m,n}=1} c_{m}(\mathbf{a}_{m,:}) \geq 1$.
Let $\underline{\mathbf{c}}(\mathbf{A}) = [\underline{c}_{1}, \underline{c}_{2}, \ldots, \underline{c}_{N}] \in \mathbb{R}^{N}$ denote the vector of the least popularity of all clients.
In particular, $\Lambda(\mathbf{A})$ satisfies $\Lambda(\mathbf{A}) = \sum_{n\in\mathcal{N}}\frac{1}{\underline{c}_{n}^{2}(\mathbf{A})}$.

To analyze the optimization of the convergence rate, we start by showing Schur-convexity and majorization.

\noindent\textbf{Lemma 5.} (Schur-convexity and majorization of $\Lambda(\mathbf{A})$)
\textit{We define $\mathbf{l}(\mathbf{A}) = \sort^{\downarrow}(\underline{\mathbf{c}}(\mathbf{A}))$ as a vector of the least popularities of all clients in a descending order.
If there exists another block activation matrix $\mathbf{A}^{\prime}$ such that $\mathbf{l}(\mathbf{A})$ is majorized by $\mathbf{l}(\mathbf{A}^{\prime})$, we have $\Lambda(\mathbf{A}) < \Lambda(\mathbf{A}^{\prime})$.}
\begin{proof}
We start by recalling majorization and Schur-convexity.
We define two descending vectors $\mathbf{x}^{\downarrow}, \mathbf{y}^{\downarrow} \in \mathbb{R}^{N}$, where $x_{n}^{\downarrow}$ and $y_{n}^{\downarrow}$ denote the $n$-th largest value of $\mathbf{x}^{\downarrow}$ and $\mathbf{y}^{\downarrow}$, respectively.
If for any $k = 1, 2, \ldots, N$, $\mathbf{x}^{\downarrow}$ and $\mathbf{y}^{\downarrow}$ satisfy $\sum_{n=1}^{k}x_{i}^{\downarrow} \leq \sum_{n=1}^{k}y_{i}^{\downarrow}$ and $\sum_{n\in\mathcal{N}}x_{i}^{\downarrow} = \sum_{n\in\mathcal{N}}y_{i}^{\downarrow}$, we say $\mathbf{x}^{\downarrow}$ is majorized by $\mathbf{y}^{\downarrow}$, i.e., $\mathbf{x}^{\downarrow} \prec \mathbf{y}^{\downarrow}$.
Then, we introduce Schur-convexity.
We say a function $f: \mathbb{R}^{N} \rightarrow \mathbb{R}$ is Schur-convex if $\mathbf{x}^{\downarrow} \prec \mathbf{y}^{\downarrow}$, $f(\mathbf{x}^{\downarrow}) \leq f(\mathbf{y}^{\downarrow})$.
Due to the property of Schur-convexity, if $f$ is a convex function, then $\sum_{n\in\mathcal{N}}f({x}_{n}^{\downarrow})$ is Schur-convex.

In our settings, 
$\frac{1}{\underline{c}_{n}^{2}(\mathbf{A})}$ is convex and monotonically decreasing \textit{w.r.t.} $\underline{c}_{n}(\mathbf{A})$ for any $n\in\mathcal{N}$, $\Lambda(\mathbf{A}) = \sum_{n\in\mathcal{N}}\frac{1}{\underline{c}_{n}^{2}(\mathbf{A})}$ is Schur-convex.
Then, based on the definition of majorization, we have $\Lambda(\mathbf{A}) < \Lambda(\mathbf{A}^{\prime})$ if $\mathbf{l}(\mathbf{A}) \prec \mathbf{l}(\mathbf{A}^{\prime})$.
\end{proof}

Lemma 5 provides key insight into optimizing the convergence rate, which is summarized in the following theorem.

\noindent\textbf{Theorem 3.} (Dominance of the least popular blocks)
\textit{The value of $\Lambda(\mathbf{A})$ depends on  $\mathbf{l}(\mathbf{A})$.
Minimizing $\Lambda(\mathbf{A})$ is equivalent to maximizing the least popularity of all clients (i.e., $\max_{\mathbf{A}}\min_{n\in\mathcal{N}}\{\underline{c}_{n}(\mathbf{A})\}$) and then minimizing the number of clients which obtain this minimum.}   
\begin{proof}
Since $\Lambda(\mathbf{A})$ is Schur-convex in the sorted vector of least popularities $\mathbf{l}(\mathbf{A})$, Lemma 5 shows that minimizing $\Lambda(\mathbf{A})$ is achieved by flattening $\mathbf{l}(\mathbf{A})$ as much as possible.  
Flattening occurs first by maximizing the minimal value across $\underline{c}_{n}(\mathbf{A}), n\in\mathcal{N}$ (i.e., maximizing the least popularity across all clients).
Then, with that minimum fixed, we reduce the number of clients which obtain this minimum.
\end{proof}

In addition, we present the following proposition.

\noindent\textbf{Proposition 1.} ($\Lambda$-optimal multiplicity)
\textit{$\Lambda(\mathbf{A})$ depends only on the multiset of $\underline{\mathbf{c}}(\mathbf{A})$.
Moreover, $\Lambda(\mathbf{A})$ is permutation invariant to $\underline{\mathbf{c}}(\mathbf{A})$.
That is, for any permutation $\pi$, if two block activation matrices $\mathbf{A}$ and $\mathbf{A}^{\prime}$ satisfy $\underline{\mathbf{c}}(\mathbf{A}^{\prime}) = \pi(\underline{\mathbf{c}}(\mathbf{A}))$, they achieve a similar convergence rate.}

\begin{proof}
The value of $\Lambda(\mathbf{A})$ depends only on the multiset of least popularities $\{\underline{c}_{1}(\mathbf{A}), \underline{c}_{2}(\mathbf{A}), \ldots, \underline{c}_{N}(\mathbf{A})\}$.
The ordering of its elements plays no role because $\Lambda(\mathbf{A}) = \sum_{n\in\mathcal{N}}\frac{1}{\underline{c}_{n}^{2}(\mathbf{A})}$ is a symmetric sum.
Therefore, for any permutation $\pi$, replacing $\underline{\mathbf{c}}(\mathbf{A})$ with $\pi(\underline{\mathbf{c}}(\mathbf{A}))$ leaves the value of the sum unchanged.
\end{proof}

The above analysis uncovers a counterintuitive insight, which is summarized in the following remark.

\noindent\textbf{Remark 1.}
\textit{While the total number of activated blocks affects the convergence rate, the factor that controls the convergence is how the least popularities are distributed across clients.}

In general, increasing the total number of activated blocks across all clients enhances the convergence rate at the cost of higher VRAM usage. 
This reveals a fundamental trade-off between convergence performance and VRAM usage: activating more blocks accelerates optimization but demands more VRAM resources on each client.

\section{Problem Formulation and $\epsilon$-constraint Lexicographic Algorithm}
\label{sec:problem_solution}
\subsection{Problem Formulation}
\label{sec:problem_formulation}
Based on the aforementioned theoretical analysis, we aim to jointly minimize $\Lambda(\mathbf{A})$ to improve the convergence rate of ZorBA and reduce each client's VRAM usage.
We formulate the problem as follows:
\begin{subequations}
\label{opt_problem}
\begin{alignat}{4}
\mathcal{P}_1:\quad &\displaystyle{\minimize_{\mathbf{A}}} & & \quad \Lambda(\mathbf{A}), \psi_{\mathrm{tot},1}(\mathbf{a}_{1}), \psi_{\mathrm{tot},2}(\mathbf{a}_{2}), \nonumber\\
&&&\quad\cdots, \psi_{\mathrm{tot},N}(\mathbf{a}_{N})
\label{eq:optProb_1} \\
&\mathrm{subject\: to} 
& &\quad \sum_{m\in\mathcal{M}} a_{m,n} \geq 1, \quad n\in\mathcal{N}, \label{c12}\\
& & & \quad \sum_{n\in\mathcal{N}} a_{m,n} \geq 1, \quad m\in\mathcal{M}, \label{c13}\\
& & & \quad a_{m,n} \in \{0, 1\}, \quad m\in\mathcal{M}, n\in\mathcal{N}, \label{c14}
\end{alignat}
\end{subequations} 
where constraint (\ref{c12}) ensures that each client activates and updates at least one block for fine-tuning.
Constraint (\ref{c13}) ensures that each block is activated and updated by at least one client.
Constraint (\ref{c14}) ensures that the block activation decisions are binary variables.
The challenges of solving $\mathcal{P}_{1}$ are twofold.
First, $\mathcal{P}_{1}$ is an NP-hard integer programming problem.
Second, $\mathcal{P}_{1}$ contains multiple objectives which depend on the block activation decisions.
To address these challenges, we propose an $\epsilon$-constraint lexicographic algorithm to achieve a close-to-optimal solution in the following subsection.

\subsection{$\epsilon$-constraint Lexicographic Algorithm}
\label{sec:solution}
We adopt the $\epsilon$-constraint method \cite{haimes1971bicriterion} to transform the original multi-objective problem into a single-objective optimization problem, which allows us to balance the convergence performance with VRAM constraints. 
In particular, for each client $n$, we denote the VRAM usage reduction as $\epsilon_{n} = \tau\psi_{\mathrm{max}, n}$, where $\tau$ denotes the desired reduction ratio in VRAM usage.
To explore a diverse set of trade-offs, we sample $\tau$ independently from a uniform distribution as $\tau\sim\mathcal{U}[0,1]$.
Let $\boldsymbol{\epsilon} = [\epsilon_{1}, \epsilon_{2}, \ldots, \epsilon_{N}]\in\mathbb{R}^{N}$ denote the resulting VRAM reduction vector for all clients.
We generate $E$ such vectors and collect them in the set $\mathcal{E}$, which is used to evaluate the candidate solutions under different VRAM reduction scenarios across clients.
We solve $\mathcal{P}_{1}$ for each reduction vector $\boldsymbol{\epsilon}\in\mathcal{E}$.
Under a given $\boldsymbol{\epsilon}$, problem $\mathcal{P}_{1}$ is reformulated as
\begin{subequations}
\label{opt_problem_v2}
\begin{alignat}{5}
\mathcal{P}_2:\quad&\displaystyle{\minimize_{\mathbf{A}}} & & \quad \Lambda(\mathbf{A})
\label{eq:optProb_2} \\
&\mathrm{subject\: to}
& & \quad \psi_{\mathrm{tot}, n}(\mathbf{a}_{n}) = \psi_{\mathrm{max}, n} - \epsilon_{n}, \quad n\in\mathcal{N}, \label{c21}\\
& & &\quad \mathrm{constraints}\:(\textrm{\ref{c12}})\!-\!(\textrm{\ref{c14}}). \nonumber
\end{alignat}
\end{subequations}
Problem $\mathcal{P}_{2}$ is difficult to solve since matrix $\mathbf{A}$ is a binary matrix with dimension $M\times N$.
For each $\boldsymbol{\epsilon}$, solving problem $\mathcal{P}_{2}$ incurs an exponential computation complexity of up to $\mathcal{O}(2^{MN})$, which is not scalable as $M$ and $N$ grow.

To reduce the complexity, we propose a lexicographic optimization algorithm.
Based on Theorem 3, we equivalently transform $\mathcal{P}_{2}$ into two subproblems, namely $\mathcal{P}_{\mathrm{I}}$ and $\mathcal{P}_{\mathrm{II}}$.
Problem $\mathcal{P}_{\mathrm{I}}$ aims to maximize the least popularity of all clients.
Let $\gamma^{\star}$ denote the optimal objective value to $\mathcal{P}_{\mathrm{I}}$.
Problem $\mathcal{P}_{\mathrm{II}}$ is the least popularity adjustment problem.
The objective is to minimize the number of clients whose least popularity remains $\gamma^{\star}$ after additional block activation. 

\noindent\textbf{Least popularity maximization problem $\mathcal{P}_{\mathrm{I}}$:}
$\mathcal{P}_{\mathrm{I}}$ is formulated as follows:
\begin{subequations}
\label{opt_prblem_stage_1}
\begin{alignat}{5}
\mathcal{P}_{\mathrm{I}}:\quad&\displaystyle{\maximize_{\mathbf{A}}} & & \quad \min_{n\in\mathcal{N}}\{\underline{c}_{n}(\mathbf{A})\}
\label{eq:obj_stage_1} \\
&\mathrm{subject\: to}
& &\quad \mathrm{constraints}\:(\textrm{\ref{c12}})\!-\!(\textrm{\ref{c14}}), (\textrm{\ref{c21}}). \nonumber
\end{alignat}
\end{subequations}
The optimal objective value of $\mathcal{P}_{\mathrm{I}}$ is given by Theorem 4.

\noindent\textbf{Theorem 4.} (Optimal least popularity)
\textit{Let $\mathcal{B}$ denote any non-empty subset of blocks, i.e., $\mathcal{B}\subseteq\mathcal{M}$ and $\mathcal{B}\neq\varnothing$.
$\gamma^{\star}$ satisfies}
\begin{align}
    \label{eq:solution_s1}
    \gamma^{\star}  = \min\limits_{\mathcal{B}}\Bigg\{\Bigg\lfloor\frac{\sum_{n\in\mathcal{N}}\min\left\{\frac{\psi_{\mathrm{max}, n} - \psi_{\mathrm{md}} - \epsilon_{n}}{(\alpha+3K+1)BLH}, \lvert\mathcal{B}\rvert\right\}}{\lvert\mathcal{B}\rvert}\Bigg\rfloor\Bigg\}.
\end{align}
\begin{proof}
For client $n\in\mathcal{N}$, given its VRAM capacity, the maximum block activation budget is given by $\lambda_{n}^{\star} = \frac{\psi_{\mathrm{max}, n} - \psi_{\mathrm{md}} - \epsilon_{n}}{(\alpha+3K+1)BLH}$.
We consider that each block in the subset $\mathcal{B}\subseteq\mathcal{M}$ is to be activated by at least $\gamma$ clients.
Then, the total number of activated blocks in this subset $\mathcal{B}$ is $\lvert\mathcal{B}\vert\gamma$.
However, client $n\in\mathcal{N}$ can only activate at most $\min\{\lambda_{n}^{\star}, \lvert\mathcal{B}\vert\}$ blocks.
The total number of activated blocks of all clients for subset $\mathcal{B}$ is $\sum_{n\in\mathcal{N}}\min\{\lambda_{n}^{\star}, \lvert\mathcal{B}\vert\}$.
To make $\gamma$ feasible, we must have $\lvert\mathcal{B}\vert\gamma \leq \sum_{n\in\mathcal{N}}\min\{\lambda_{n}^{\star}, \lvert\mathcal{B}\vert\}$.
Note that this inequality must hold for all possible subsets $\mathcal{B}$.
Hence, the optimal $\gamma$ is 
\begin{align}
    \gamma^{\star} 
    =&\: \min\limits_{\mathcal{B}}\Big\{\gamma:  \lvert\mathcal{B}\vert\gamma \leq \sum_{n\in\mathcal{N}}\min\{\lambda_{n}^{\star}, \lvert\mathcal{B}\vert\}, \mathcal{B}\subseteq\mathcal{M}, \mathcal{B}\neq\varnothing\Big\} \nonumber\\
    =&\: \min\limits_{\mathcal{B}}\bigg\{\bigg\lfloor\frac{\sum_{n\in\mathcal{N}}\min\left\{\frac{\psi_{\mathrm{max}, n} - \psi_{\mathrm{md}} - \epsilon_{n}}{(\alpha+3K+1)BLH}, \lvert\mathcal{B}\rvert\right\}}{\lvert\mathcal{B}\rvert}\bigg\rfloor\bigg\}.
\end{align}
\end{proof}

After $\gamma^{\star}$ has been determined, we apply Dinic's algorithm \cite{dinitz2006dinitz} to construct the optimal block activation matrix for problem $\mathcal{P}_{\mathrm{I}}$.
We denote this matrix as $\Tilde{\mathbf{A}} \in \{0,1\}^{M \times N}$, which serves as the initial block activation matrix.
Based on $\Tilde{\mathbf{A}}$, we obtain the initial popularity of each block and initial least popularity of each client as $\Tilde{c}_{m}, m\in\mathcal{M}$ and $\Tilde{\underline{c}}_{n}, n\in\mathcal{N}$, respectively.
The initial popularity vector and initial least popularity vector are denoted as $\Tilde{\mathbf{c}}\in\mathbb{R}^{M}$ and $\Tilde{\underline{\mathbf{c}}}\in\mathbb{R}^{N}$, respectively.

\noindent\textbf{Least popularity adjustment problem $\mathcal{P}_{\mathrm{II}}$:} Note that $\Tilde{\mathbf{A}}$ may not be optimal to problem $\mathcal{P}_{2}$.
Hence, based on Theorem 3, we formulate problem $\mathcal{P}_{\mathrm{II}}$ to activate as many additional blocks as possible on top of $\Tilde{\mathbf{A}}$ such that the number of clients with the least popularity $\gamma^{\star}$ is minimized.
Let $x_{m,n}\in\{0,1\}$ denote the additional block activation decision.
We set $x_{m,n} = 1$ to additionally activate the $m$-th block of client $n$ when $\Tilde{a}_{m,n} = 0$.
Otherwise,  $x_{m,n} = 0$.
That is,
\begin{align}
\label{eq:x_constraint}
    x_{m,n}\Tilde{a}_{m,n} = 0, \quad m\in\mathcal{M}, n\in\mathcal{N}.
\end{align}

Let $\mathbf{X}\in\{0,1\}^{M\times N}$ denote the additional block activation decision matrix.
Note that the total number of blocks that each client activates must satisfy the VRAM constraint.
That is, 
\begin{align}
    \label{eq:VRAM_constraint_II}
    \sum_{m\in\mathcal{M}}(x_{m,n} + \Tilde{a}_{m,n})\leq \lambda_{n}^{\star}, \quad n\in\mathcal{N}.
\end{align}

Then, given the decision matrix $\mathbf{X}$, we obtain the popularity of each block and the least popularity of each client as
\begin{align}
    c_{m} &= \Tilde{c}_{m} + \sum_{n\in\mathcal{N}}x_{m,n}, \quad m\in\mathcal{M}, \label{eq:c_m_contraint}\\
    \underline{c}_{n} &= \min\limits_{m:\Tilde{a}_{m,n} + x_{m,n} =1} c_{m}, \quad n\in\mathcal{N}. \label{eq:c_n_constraint}
\end{align}

In addition, we introduce an auxiliary indicator variable $y_n \in \{0, 1\}$ to characterize whether client $n$'s least popularity is still $\gamma^{\star}$ after additional block activation.
In particular, if $\underline{c}_{n} = \gamma^{\star}$, then $y_{n} = 1$.
Otherwise, $y_{n} = 0$.
If client $n$ activates the $m$-th block after additional block activation and its least popularity is larger than $\gamma^{\star}$, then the popularity of the $m$-th block must be larger than $\gamma^{\star}$.
That is, for $x_{m,n} + \Tilde{a}_{m,n} = 1$ and $y_{n}=0$, $c_{m}$ must satisfy $c_{m} \geq \gamma^{\star} + 1, m\in\mathcal{M}, n\in\mathcal{N}$.
We use the big-M method \cite{griva2008linear} to linearize this condition as
\begin{align}
    \label{eq:big-M-v1}
    c_{m} \geq (\gamma^{\star}+1) - \beta_{1}\left(1-(x_{m,n} + \Tilde{a}_{m,n})\right) - \beta_{2}y_{n},& \nonumber\\
    m\in\mathcal{M}, n\in\mathcal{N},&
\end{align}
where $\beta_{1}$ and $\beta_{2}$ are two positive constants.
When $\beta_{1} = \beta_{2} = \gamma^{\star}+1$, the aforementioned condition always holds. 
Thus, inequality (\ref{eq:big-M-v1}) is equivalent to
\begin{align}
    \label{eq:big-M-v2}
    c_{m} \
    \geq&\: (\gamma^{\star}+1) - (\gamma^{\star}+1)\left(1-(x_{m,n} + \Tilde{a}_{m,n})\right) \nonumber\\
    &- (\gamma^{\star}+1)y_{n}, \quad m\in\mathcal{M}, n\in\mathcal{N}.
\end{align}

In summary, problem $\mathcal{P}_{\mathrm{II}}$ can be formulated as follows:
\begin{subequations}
\label{opt_prblem_stage_2}
\begin{alignat}{5}
\mathcal{P}_{\mathrm{II}}:\quad&\displaystyle{\minimize_{\mathbf{X}, \{y_{n}\}_{n\in\mathcal{N}}}} & & \quad \sum_{n\in\mathcal{N}}y_{n}
\label{eq:obj_s2} \\
&\mathrm{subject\: to}
& & \quad \mathrm{constraints }\:(\ref{eq:x_constraint})\!-\!(\ref{eq:c_n_constraint}),\: (\ref{eq:big-M-v2}), \nonumber\\
& & &\quad x_{m,n}, y_{n}\in\{0,1\}, \quad m\in\mathcal{M}, n\in\mathcal{N}.
\end{alignat}
\end{subequations}
To solve problem $\mathcal{P}_{\mathrm{II}}$, we propose a greedy update algorithm.
We start by introducing several terms.
Let $\mathcal{L} = \{m: \Tilde{c}_{m} = \gamma^{\star}, m\in\mathcal{M}\}$ denote the set of blocks with its initial popularity as $\gamma^{\star}$.
Let $\mathcal{F} = \{n: \Tilde{\underline{c}}_{n} = \gamma^{\star}, n\in\mathcal{N}\}$ denote the set of clients whose initial least popularity is $\gamma^{\star}$.
Let $\mathbf{r} \in \mathbb{R}^{N}$ denote the vector of remaining VRAM budgets, in which each element $r_{n} = \lfloor\lambda_n^{\star}\rfloor - \sum_{m \in \mathcal{M}} \tilde{a}_{m,n}$ represents the number of additional blocks client $n$ can activate after the initial block activation.
For client $n\in\mathcal{N}$, we denote $\mathcal{W}_{n} = \{m\in\mathcal{L}: \Tilde{a}_{m,n} = 1\}$ as the ``bottleneck" set.
It includes those blocks which are activated while satisfying the initial popularity.

The greedy update algorithm can be summarized as follows:
We define $\mathrm{Gain}\_{\mathrm{ls}}[m]$ as the number of clients whose current least popularity is $\gamma^{\star}$ and would increase to $\gamma^{\star} + 1$ if their $m$-th block were activated. 
Among all blocks with popularity $\gamma^{\star}$ (i.e., $m \in \mathcal{L}$), we denote $m^{\star}$ as the ``most valuable" block that maximally reduces the number of clients with least popularity $\gamma^{\star}$ when activated. 
We then activate the $m^{\star}$-th block by assigning it to any client $n$ with remaining VRAM budget to activate more blocks (i.e., $r_{n} > 0$) and has not yet activated $m^{\star}$-th block (i.e., $\tilde{a}_{m^{\star},n} = 0$). 
Thus, the popularity of the $m^{\star}$-th block increases from $\gamma^{\star}$ to $\gamma^{\star} + 1$, i.e., $c_{m^{\star}} \leftarrow \gamma^{\star} + 1$.
Next, for every client $n \in \mathcal{N}$ whose ``bottleneck" set $\mathcal{W}_n$ includes block $m^{\star}$, we remove $m^{\star}$ from $\mathcal{W}_n$. 
If a client's bottleneck set becomes empty, it is removed from the set of remaining clients $\mathcal{F}$. 
This procedure is repeated iteratively until the number of clients with the least popularity $\gamma^{\star}$ is minimized. 
We summarize this algorithm in Algorithm \ref{alg:greedy_alg}.

To solve problem $\mathcal{P}_{1}$, we iterate over each $\boldsymbol{\epsilon} \in \mathcal{E}$.
For each $\boldsymbol{\epsilon}$, we solve the corresponding subproblems $\mathcal{P}_{\mathrm{I}}$ and $\mathcal{P}_{\mathrm{II}}$, and obtain the optimal block activation matrix $\mathbf{A}_{\boldsymbol{\epsilon}}^{\star}$.
Based on $\mathbf{A}_{\boldsymbol{\epsilon}}^{\star}$, we can determine the corresponding value of $\Lambda(\mathbf{A}_{\boldsymbol{\epsilon}}^{\star})$ and $\psi_{\mathrm{tot},n}(\mathbf{A}_{\boldsymbol{\epsilon}}^{\star}), n\in\mathcal{N}$.
We denote the total usage of all clients as $\sum_{n\in\mathcal{N}}\psi_{\mathrm{tot},n}(\mathbf{A}_{\boldsymbol{\epsilon}}^{\star})$.
By going over all $\boldsymbol{\epsilon}\in\mathcal{E}$, we can form a Pareto front which characterizes the trade-off between $\Lambda$ and total VRAM usage of all clients.
We can choose the optimal block activation matrix $\mathbf{A}^{\star}$ on the Pareto front to balance the VRAM usage and the convergence rate for ZorBA.
The algorithm to solve problem $\mathcal{P}_{1}$ is shown in Algorithm \ref{alg:prob_solver}.

\begin{algorithm}[t] 
\small
\caption{Greedy update algorithm for $\mathcal{P}_{\mathrm{II}}$}
\label{alg:greedy_alg}   
\begin{algorithmic}[1] 
    \STATE \textbf{Input}: Initial block activation matrix $\Tilde{\mathbf{A}}$; 
    initial popularity vector $\Tilde{\mathbf{c}}$; 
    initial least popularity vector $\Tilde{\underline{\mathbf{c}}}$;
    optimal least popularity $\gamma^{\star}$;
    $\mathcal{W}_{n}, n\in\mathcal{N}$;
    $\mathcal{L}$; 
    $\mathcal{F}$;
    $\mathbf{r}$;
    $\lambda_{n}^{\star}, n\in\mathcal{N}$;
    $\mathrm{Gain}\_{\mathrm{ls}}$ = [].
    
    \STATE \textbf{While} $\mathcal{L} \neq \varnothing$, $\mathcal{F} \neq \varnothing$, and $\sum_{n\in\mathcal{N}}\lambda_{n}^{\star} > 0$ \textbf{do}
    \begin{ALC@g}

    \STATE Create a candidate block index set $\mathcal{C} := \{m\in\mathcal{L}: \exists n\in\mathcal{N}, \:\mathrm{such\: that}\: \tilde{a}_{m,n} = 0\:\mathrm{and}\:r_{n} > 0 \}$.


    \STATE \textbf{For} $m \in \mathcal{C}$ \textbf{do}
    \begin{ALC@g}

    \STATE $\mathrm{Gain}\_{\mathrm{ls}}$[$m$] := $\vert\{n\in\mathcal{F}:\:m\in\mathcal{W}_{n}\} \vert$.
    
    \end{ALC@g}
    \STATE \textbf{End for}

    \STATE $m^{\star} := \argmax_{m\in\mathcal{C}}\{\mathrm{Gain}\_{\mathrm{ls}}[m]\}$.

    \STATE Randomly select client $n$ such that $r_{n} > 0$ and $\Tilde{a}_{m^{\star}, n} = 0$.

    \STATE $x_{m^{\star}, n} := 1$, $r_{n} := r_{n} - 1$. 
    
    \STATE Remove $m^{\star}$ from $\mathcal{L}$.

    \STATE \textbf{For} $n \in \mathcal{N}$ and $\Tilde{a}_{m^{\star}, n} = 1$ \textbf{do}
    \begin{ALC@g}

    \STATE Remove $m^{\star}$ from $\mathcal{W}_{n}$.
    Remove $n$ from $\mathcal{F}$ if $\mathcal{W}_{n} = \varnothing$.
    
    \end{ALC@g}
    \STATE \textbf{End for}

    \end{ALC@g}
    \STATE \textbf{End while}

    \STATE $\mathbf{A} := \Tilde{\mathbf{A}} + \mathbf{X}$.

    \STATE \textbf{Output}: Updated block activation matrix $\mathbf{A}$.
\end{algorithmic}
\end{algorithm}

By using Dinic's algorithm, solving problem $\mathcal{P}_{\mathrm{I}}$ incurs a computation complexity of $\mathcal{O}(MN\sqrt{M+N}\log_{2}N)$.
The greedy update algorithm in the second stage incurs a computation complexity of $\mathcal{O}(\min\{M, N\}MN)$.
By going through $\mathcal{E}$, our proposed $\epsilon$-constraint lexicographic algorithm incurs a total computation complexity of $\mathcal{O}(EMN(\sqrt{M+N}\log_{2}N + \min\{M, N\}))$, which is significantly lower than that of directly solving problem $\mathcal{P}_{1}$.
The overall workflow of our proposed ZorBA is shown in Algorithm \ref{alg:ZorBA}.
\begin{algorithm}[t] 
\small
\caption{$\epsilon$-constraint lexicographic algorithm for $\mathcal{P}_{1}$}
\label{alg:prob_solver}   
\begin{algorithmic}[1] 
    \STATE \textbf{Input}: VRAM capacities $\psi_{\max, n}$, $n\in\mathcal{N}$; initial constraint parameter vector $\mathcal{E}$.

    \STATE \textbf{For} $\boldsymbol{\epsilon} \in \mathcal{E}$ \textbf{do}
    \begin{ALC@g}

    \STATE Solve problem $\mathcal{P}_{\mathrm{I}}$ to determine $\gamma^{\star}$ by using eqn. (\ref{eq:solution_s1}).

    \STATE Construct the initial block activation matrix $\Tilde{\mathbf{A}}$.

    \STATE Obtain the updated block activation matrix $\mathbf{A}_{\boldsymbol{\epsilon}}^{\star}$ using Alg. \ref{alg:greedy_alg}.

    \STATE Determine $\Lambda(\mathbf{A}_{\boldsymbol{\epsilon}}^{\star})$ and $\psi_{\mathrm{tot},n}(\mathbf{A}_{\boldsymbol{\epsilon}}^{\star}), n\in\mathcal{N}$.

    \STATE Record tuple ($\boldsymbol{\epsilon}$, $\mathbf{A}_{\boldsymbol{\epsilon}}^{\star}$, $\Lambda(\mathbf{A}_{\boldsymbol{\epsilon}}^{\star})$, $\sum_{n\in\mathcal{N}}\psi_{\mathrm{tot},n}(\mathbf{A}_{\boldsymbol{\epsilon}}^{\star})$).

    \end{ALC@g}
    \STATE \textbf{End for}

    \STATE Select $\mathbf{A}^{\star}$ on the Pareto front $\{\sum_{n\in\mathcal{N}}\psi_{\mathrm{tot},n}(\mathbf{A}_{\boldsymbol{\epsilon}}^{\star}), \Lambda(\mathbf{A}_{\boldsymbol{\epsilon}}^{\star})\}_{\boldsymbol{\epsilon}\in\mathcal
    E}$.

    \STATE \textbf{Output}: Optimal block activation matrix $\mathbf{A}^{\star}$.
\end{algorithmic}
\end{algorithm}

\begin{algorithm}[t] 
\small
\caption{ZorBA Algorithm}
\label{alg:ZorBA}   
\begin{algorithmic}[1] 
    \STATE \textbf{Input}: Initial model $\Bar{\mathbf{w}}^{1}$; local fine-tuning datasets $\mathcal{D}_{n}$, $n\in\mathcal{N}$; initial random seeds $\mathcal{S}$; learning rate $\eta$.

    \STATE The central server solves problem $\mathcal{P}_{1}$ to determine the block activation matrix $\mathbf{A}^{\star}$ by using Algorithm \ref{alg:prob_solver}. 

    \STATE \textbf{For} client $n \in \mathcal{N}$ \textbf{in parallel do}
    \begin{ALC@g}
    
        \STATE Activate local model parameters.

    \end{ALC@g}
    \STATE \textbf{End for}

    \STATE \textbf{For} $t \in \mathcal{T}$ \textbf{do}
    \begin{ALC@g}

    \STATE The central server selects a subset $\mathcal{Q}^{t}$ from $\mathcal{S}$.

    \STATE \textbf{For} client $n \in \mathcal{N}$ \textbf{in parallel do}
    \begin{ALC@g}
            
        \STATE Estimate the gradients $\Tilde{\nabla} F_{n}(\Hat{\mathbf{w}}_{n}^{t}; \bm{\xi}_{n})$ based on eqn. (\ref{eq:zo_est_seed}) and record the finite differences of the estimated gradients.


        \STATE Transmit the finite differences of the estimated gradients to the central server.
        
        \end{ALC@g}
        \STATE \textbf{End for}

    \STATE The central server performs aggregation based on eqn. (\ref{eq:model_agg}), obtains and broadcasts $\mathcal{\bar{P}}^{t}$ back to the clients.

    \STATE \textbf{For} client $n \in \mathcal{N}$ \textbf{in parallel do}
    \begin{ALC@g}
    
        \STATE Determine $\Bar{\mathbf{w}}^{t+1}$ based on $\mathcal{\bar{P}}^{t}$.

    \end{ALC@g}
    \STATE \textbf{End for}
        
    \end{ALC@g}
    \STATE \textbf{End for}

    \STATE \textbf{Output}: fine-tuned model $\Bar{\mathbf{w}}^{T+1}$.
\end{algorithmic}
\end{algorithm}

\section{Performance Evaluation}
\label{sec:performance_evaluation}
\subsection{Simulation Setup}
\label{sec:simulation_setup}
We conduct federated LLM fine-tuning over $N = 50$ clients and a central server.
We use OPT-125M and OPT-1.3B \cite{zhang2022opt} as the local models for experiments.
In particular, OPT-125M and OPT-1.3B have 12 and 24 transformer blocks, respectively.
We conduct experiments on text classification datasets, including AG-News \cite{AG-news}, SST-2 \cite{socher2013recursive}, and SNLI \cite{bowman2015large}.
We use the Dirichlet distribution $\mathrm{Dir}(1.0)$ to create non-independent and identically distributed data partitioning across clients' local datasets.
To characterize the VRAM capacity, we denote the maximum VRAM of all clients as the sum of the number of model parameters and activations of all blocks.
We set the minimum VRAM of all clients to the sum of the number of model parameters and activations of one block. 
The VRAM capacity of each client follows a uniform distribution between the minimum and maximum of VRAM.
We use the number of transmitted parameters to characterize the communication overhead.
In addition, we set $\alpha = 4$, $\eta = 5\times 10^{-5}$, $\mu=10^{-4}$, $B = 8$, $E = 1000$, $P = 4096$, $Q = 10$, and $T = 500$.
We compare the performance of ZorBA with the following baseline schemes:
\begin{itemize}
    \item FedIT \cite{10447454}:
    Clients perform first-order BP with all blocks. 

    \item FedZO \cite{9917343}:
    Clients sample one perturbation vector to perform zeroth-order optimization with all blocks during each fine-tuning round.
    Then, clients exchange the estimated gradients with the central server.

    \item DeComFL \cite{li2024achieving}:
    Clients use a shared perturbation vector to perform zeroth-order optimization with all blocks during each fine-tuning round. 
    Then, clients exchange the finite differences of estimated gradients with the server.

\end{itemize}

\subsection{Experiments}
\label{sec:experiments}
\subsubsection{Comparison of the Convergence}
To show the convergence rate, we compare the number of training rounds each approach takes to achieve a target accuracy.
In particular, for AG-News, SST-2, and SNLI, we set the target accuracies as 0.8, 0.8, and 0.72, respectively.
Since the central server determines the block activation matrix in prior, we present the total VRAM usage of all clients per fine-tuning round, which remains consistent during fine-tuning.
Furthermore, we use the number of parameters transmitted between clients and the central server to characterize the total communication overhead.
Results in Table \ref{tb:acc} show that our proposed ZorBA converges faster than DeComFL by up to 23.76\%.
ZorBA also converges faster than FedZO in most cases.
These results highlight the benefit of optimizing the heterogeneous block activation matrix to accelerate the convergence.

\subsubsection{Comparison of the Total VRAM Usage and Communication Overhead}
We compare the total VRAM usage and communication overhead in Table \ref{tb:vram_commun}.
Our proposed ZorBA achieves a much lower total VRAM usage, which is up to 62.41\%, 54.75\%, and 54.75\% lower than FedIT, FedZO, and DeComFL, respectively.
Note that FedZO and DeComFL incur the same total VRAM usage since they both activate all blocks.
In addition, ZorBA incurs negligible communication overhead when compared with FedIT and FedZO. 
ZorBA also achieves a comparable total communication overhead to DeComFL.

\begin{table}[t]
\centering
\caption{Comparison of the number of fine-tuning rounds required to achieve the target average testing accuracy.}
\label{tb:acc}
\resizebox{0.38\textwidth}{!}{
\begin{tabular}{c|c|cccc}
\specialrule{1.5pt}{0pt}{0pt}
\textbf{Model} & \textbf{Datasets}  & \textbf{ZorBA} &
\textbf{FedIT} & \textbf{FedZO} & \textbf{DeComFL}\\
\specialrule{1.5pt}{0pt}{0pt}
\multirow{3}{*}{OPT-125M} & AG-News & 138 & 49 & 155 & 181\\
\cline{2-6}
&SST-2 & 167 & 70 & 161 & 205\\
\cline{2-6}
&SNLI  & 231 & 112 & 255 & 246\\
\specialrule{1.5pt}{0pt}{0pt}
\multirow{3}{*}{OPT-1.3B} & AG-News & 61 & 23 & 79 & 68\\
\cline{2-6}
&SST-2 & 99 & 48 & 104 & 104\\
\cline{2-6}
&SNLI  & 157 & 82 & 186 & 173\\
\specialrule{1.5pt}{0pt}{0pt}
\end{tabular}}
\end{table}

\begin{table}[t]
\centering
\caption{Comparison of the total VRAM usage of all clients per round (in GB) and total communication overhead (i.e., Commun.) of all clients to achieve the target accuracy.}
\label{tb:vram_commun}
\resizebox{0.48\textwidth}{!}{
\begin{tabular}{c|c|c|cccc}
\specialrule{1.5pt}{0pt}{0pt}
\textbf{Model} & \textbf{Datasets} & \textbf{Metrics} & \textbf{ZorBA} &
\textbf{FedIT} & \textbf{FedZO} & \textbf{DeComFL}\\
\specialrule{1.5pt}{0pt}{0pt}
\multirow{6}{*}{OPT-125M} &\multirow{2}{*}{AG-News}& VRAM  & 47.31 & 118.67   & 95.37 & 95.37  \\
& & \shortstack{Commun. } & $1.24\times 10^{5}$ & $6.13\times 10^{11}$ & $1.94\times 10^{12}$ & $1.36\times 10^{5}$\\
\cline{2-7}
&\multirow{2}{*}{SST-2}& VRAM  & 59.32 & 118.67  & 95.37  & 95.37  \\
& & \shortstack{Commun. } & $1.67\times 10^{5}$ & $8.75\times10^{11}$& $2.01\times10^{12}$& $1.54\times 10^{5}$\\
\cline{2-7}
&\multirow{2}{*}{SNLI}& VRAM  & 131.39 & 190.74  & 167.44  & 167.44 \\
& & \shortstack{Commun. } & $2.43\times 10^{5}$ & $1.40\times10^{12}$ & $3.19\times10^{12}$ & $1.85\times 10^{5}$\\
\specialrule{1.5pt}{0pt}{0pt}
\multirow{6}{*}{OPT-1.3B} &\multirow{2}{*}{AG-News}& VRAM  & 538.04 & 1431.17  & 1189.03 & 1189.03 \\
& & \shortstack{Commun. } & $7.93\times 10^{4}$ & $2.99\times10^{12}$ & $1.03\times10^{13}$ & $9.18\times10^{4}$\\
\cline{2-7}
&\multirow{2}{*}{SST-2}& VRAM  & 794.49 & 1431.17  & 1189.03  & 1189.03 \\
& & \shortstack{Commun. } & $1.34\times 10^{5}$ & $6.24\times10^{12}$ & $1.35\times10^{13}$ & $1.40\times10^{5}$\\
\cline{2-7}
&\multirow{2}{*}{SNLI}& VRAM  & 1583.55 & 2378.05 & 2315.90 & 2315.90\\
& & \shortstack{Commun. } & $2.27\times 10^{5}$ & $1.07\times10^{13}$ & $2.42\times10^{13}$ & $2.34\times10^{5}$\\
\specialrule{1.5pt}{0pt}{0pt}
\end{tabular}}
\end{table}

\begin{figure}[t]
    \centering
    
    \subfigure[]{\raisebox{-1.3mm}{\includegraphics[width=0.22\textwidth]{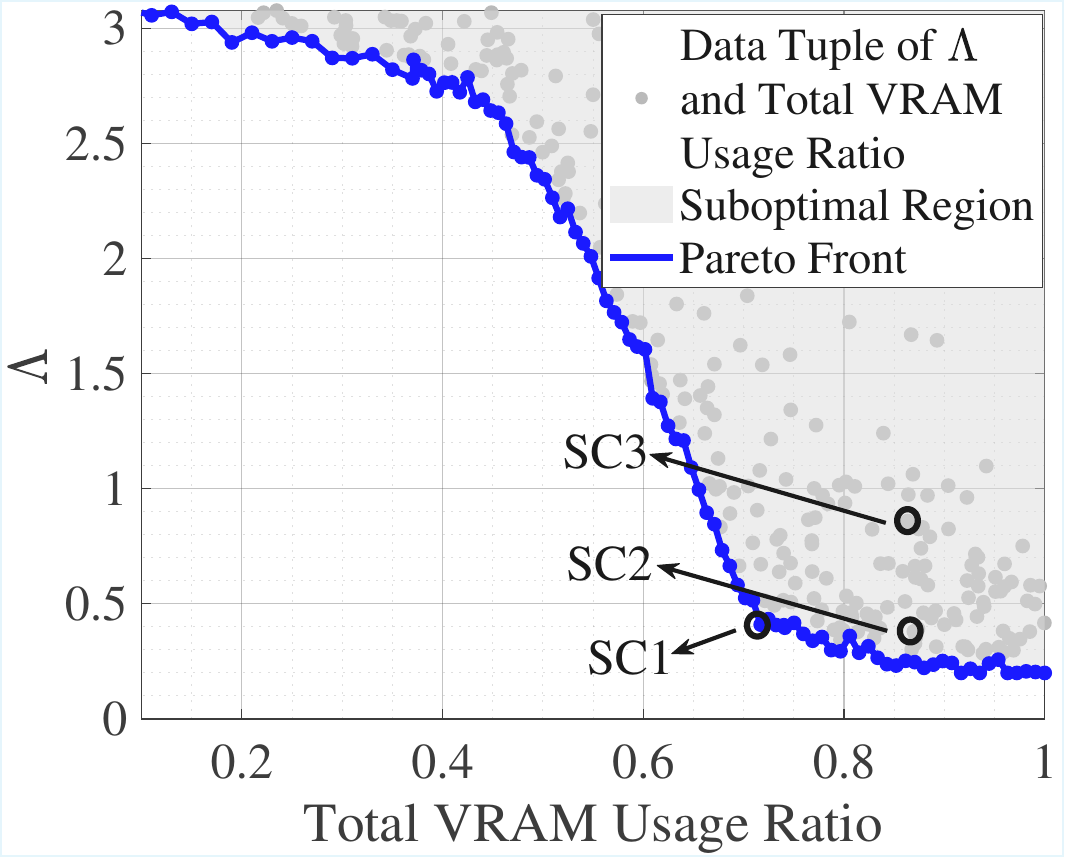}}} 
    \subfigure[]{\includegraphics[width=0.24\textwidth]{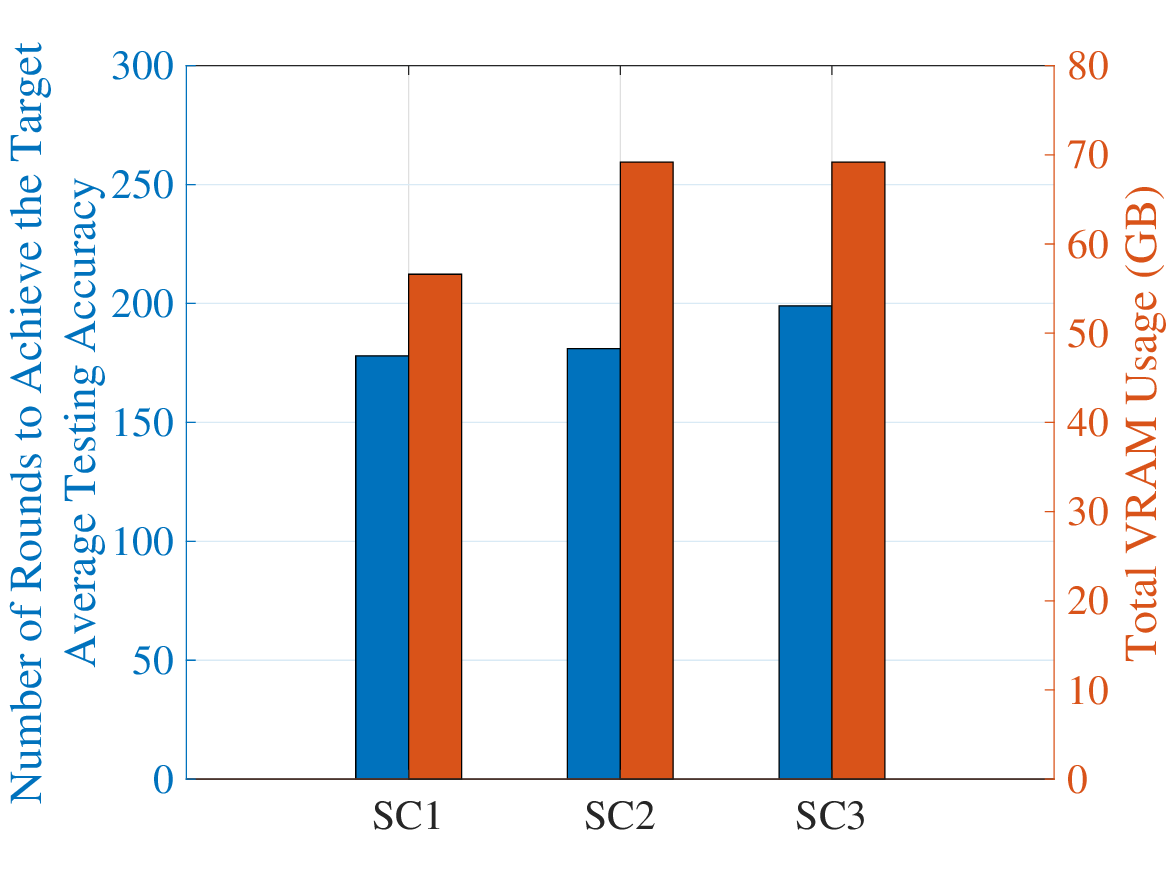}} 
    \caption{(a) Pareto front of $\Lambda$ versus the total VRAM usage ratio. (b) Comparison between SC1, SC2, and SC3 for the number of rounds to achieve the target average testing accuracy and the total VRAM usage.}
    \label{fig:case_study}
\end{figure}

\subsubsection{Trade-off between $\Lambda$ and Total VRAM Usage}
To show the trade-off between the value of $\Lambda$ and the total VRAM usage of all clients, we conduct experiments by using OPT-125M and SST-2 dataset.
For better illustration, we use the fraction of the total VRAM usage of all clients (i.e., $\frac{\sum_{n\in\mathcal{N}}\psi_{\mathrm{tot},n}}{\sum_{n\in\mathcal{N}}\psi_{\max,n}}$).
The Pareto front is shown in Fig. \ref{fig:case_study} (a).
Results show that by activating more blocks across clients, $\Lambda$ generally exhibits a descending trend, which substantiates our Observation 1 in Section \ref{sec:theoretical_analysis}.
Moreover, the value of $\Lambda$ has a trend of initially slow decline, followed by a sharp drop in the middle, and gradually converging to the minimum.
Based on this trend, we can empirically determine the block activation matrix such that the total VRAM usage is effectively reduced without degrading the convergence rate of our proposed ZorBA.

\subsubsection{Case Study}
To further demonstrate the relationship between $\Lambda$ and the convergence rate of our proposed ZorBA, we conduct three case studies.
In particular, we select three scenarios from Fig. \ref{fig:case_study} (a), namely SC1, SC2, and SC3, respectively.
The values of the data tuples of SC1, SC2, and SC3 are (0.71, 0.50), (0.86, 0.50), and (0.86, 0.95), respectively.
The comparison of the number of fine-tuning rounds to achieve the target average testing accuracy of 0.8 and the total VRAM usage of all clients is shown in Fig. \ref{fig:case_study} (b).
Results show that SC1 requires less total VRAM than SC2, indicating fewer activated blocks across all clients. Nevertheless, SC1 achieves a similar (a bit faster) convergence rate because both SC1 and SC2 yield comparable 
values of $\Lambda$. 
This supports Observation 2 and Proposition 1 in Section \ref{sec:theoretical_analysis}.
In contrast, SC3 uses more VRAM than SC1 and matches SC2 in the total number of activated blocks.
However, it needs more fine-tuning rounds to reach the target accuracy.
This slower convergence highlights the role of $\Lambda$, which validates Observations 3 and 4.

\section{Conclusion}
\label{sec:conclusion}
In this work, we proposed ZorBA, a zeroth-order optimization-based federated fine-tuning framework.
ZorBA leverages zeroth-order optimization and a heterogeneous block activation mechanism to reduce the VRAM usage.
It introduces shared random seeds to reduce the communication overhead.
We theoretically analyzed the convergence rate of our proposed ZorBA and investigated the impact of block activation decisions on the convergence rate and VRAM usage.
We proposed an $\epsilon$-constraint lexicographic algorithm to optimize the block activation decisions.
Experimental results show that ZorBA converges faster than zeroth-order optimization baselines, while reducing VRAM usage by up to 62.41\% and incurring negligible communication overhead.

\newpage
\bibliography{ieeebib}

@article{zhao2023survey,
  title={A survey of large language models},
  author={Zhao, Wayne Xin and Zhou, Kun and Li, Junyi and Tang, Tianyi and Wang, Xiaolei and Hou, Yupeng and Min, Yingqian and Zhang, Beichen and Zhang, Junjie and Dong, Zican and others},
  journal={arXiv preprint arXiv:2303.18223},
  year={Mar. 2023}
}

@article{achiam2023gpt,
  title={{GPT}-4 technical report},
  author={Achiam, Josh and Adler, Steven and Agarwal, Sandhini and Ahmad, Lama and Akkaya, Ilge and Aleman, Florencia Leoni and Almeida, Diogo and Altenschmidt, Janko and Altman, Sam and Anadkat, Shyamal and others},
  journal={arXiv preprint arXiv:2303.08774},
  year={Mar. 2023}
}

@article{grattafiori2024llama,
  title={The {L}lama 3 herd of models},
  author={Grattafiori, Aaron and Dubey, Abhimanyu and Jauhri, Abhinav and Pandey, Abhinav and Kadian, Abhishek and Al-Dahle, Ahmad and Letman, Aiesha and Mathur, Akhil and Schelten, Alan and Vaughan, Alex and others},
  journal={arXiv preprint arXiv:2407.21783},
  year={Jul. 2024}
}

@article{team2024gemma,
  title={Gemma 2: Improving open language models at a practical size},
  author={Team, Gemma and Riviere, Morgane and Pathak, Shreya and Sessa, Pier Giuseppe and Hardin, Cassidy and Bhupatiraju, Surya and Hussenot, L{\'e}onard and Mesnard, Thomas and Shahriari, Bobak and Ram{\'e}, Alexandre and others},
  journal={arXiv preprint arXiv:2408.00118},
  year={Jul. 2024}
}

@INPROCEEDINGS{11044734,
  author={Ye, Shengyuan and Ouyang, Bei and Zeng, Liekang and Qian, Tianyi and Chu, Xiaowen and Tang, Jian and Chen, Xu},
  booktitle={Proc. IEEE INFOCOM}, 
  title={Jupiter: Fast and Resource-Efficient Collaborative Inference of Generative {LLM}s on Edge Devices}, 
  year={London, UK, May 2025},
  volume={},
  number={},
  pages={}}

@article{ding2023parameter,
  title={Parameter-efficient fine-tuning of large-scale pre-trained language models},
  author={Ding, Ning and Qin, Yujia and Yang, Guang and Wei, Fuchao and Yang, Zonghan and Su, Yusheng and Hu, Shengding and Chen, Yulin and Chan, Chi-Min and Chen, Weize and others},
  journal={Nature Mach. Intell.},
  volume={5},
  number={3},
  pages={220--235},
  year={Mar. 2023},
}

@inproceedings{hu2022lora,
  title={Lo{RA}: Low-rank adaptation of large language models},
  author={Hu, Edward J and Shen, Yelong and Wallis, Phillip and Allen-Zhu, Zeyuan and Li, Yuanzhi and Wang, Shean and Wang, Lu and Chen, Weizhu and others},
  booktitle={Proc. Int. Conf. Learn. Representations (ICLR)},
  year={May 2022}
}

@inproceedings{liu2022few,
  title={Few-shot parameter-efficient fine-tuning is better and cheaper than in-context learning},
  author={Liu, Haokun and Tam, Derek and Muqeeth, Mohammed and Mohta, Jay and Huang, Tenghao and Bansal, Mohit and Raffel, Colin A},
  booktitle={Proc. Advances Neural Info. Process. Syst. (NeurIPS)},
  year={Dec. 2022}
}

@inproceedings{mcmahan2017communication,
  title={Communication-Efficient learning of deep networks from decentralized data},
  author={McMahan, Brendan and Moore, Eider and Ramage, Daniel and Hampson, Seth and y Arcas, Blaise Aguera},
  booktitle={Proc. Int. Conf. Artif. Intell. Statist. (AISTATS)},
  year={Fort Lauderdale, FL, Apr. 2017}
}

@inproceedings{li2020federated,
  title={Federated optimization in heterogeneous networks},
  author={Li, Tian and Sahu, Anit Kumar and Zaheer, Manzil and Sanjabi, Maziar and Talwalkar, Ameet and Smith, Virginia},
  booktitle={Proc. Mach. Learn. Syst. (MLSys)},
  year={Mar. 2020}
}

@inproceedings{karimireddy2020scaffold,
  title={{SCAFFOLD}: Stochastic controlled averaging for federated learning},
  author={Karimireddy, Sai Praneeth and Kale, Satyen and Mohri, Mehryar and Reddi, Sashank and Stich, Sebastian and Suresh, Ananda Theertha},
  booktitle={Proc. Int. Conf. Mach. Learn. (ICML)},
  year={Jul. 2020},
}

@inproceedings{wang2020tackling,
  title={Tackling the objective inconsistency problem in heterogeneous federated optimization},
  author={Wang, Jianyu and Liu, Qinghua and Liang, Hao and Joshi, Gauri and Poor, H Vincent},
  booktitle={Advances Neural Info. Process. Syst. (NeurIPS)},
  year={Dec. 2020}
}

@INPROCEEDINGS{10447454,
  author={Zhang, Jianyi and Vahidian, Saeed and Kuo, Martin and Li, Chunyuan and Zhang, Ruiyi and Yu, Tong and Wang, Guoyin and Chen, Yiran},
  booktitle={Proc. IEEE Int. Conf. Acoust. Speech Signal Process. (ICASSP)}, 
  title={Towards Building The Federated {GPT}: Federated Instruction Tuning}, 
  year={Seoul, Korea, Apr. 2024},
}

@inproceedings{wang2024flora,
  title={{FL}o{RA}: Federated fine-tuning large language models with heterogeneous low-rank adaptations},
  author={Wang, Ziyao and Shen, Zheyu and He, Yexiao and Sun, Guoheng and Wang, Hongyi and Lyu, Lingjuan and Li, Ang},
  booktitle={Proc. Advances Neural Info. Process. Syst. (NeurIPS)},
  year={Vancouver, Canada, Dec. 2024}
}

@inproceedings{bai2024federated,
  title={Federated fine-tuning of large language models under heterogeneous tasks and client resources},
  author={Bai, Jiamu and Chen, Daoyuan and Qian, Bingchen and Yao, Liuyi and Li, Yaliang},
  booktitle={Proc. Advances Neural Info. Process. Syst. (NeurIPS)},
  year={Vancouver, Canada, Dec. 2024}
}

@ARTICLE{10855336,
  author={Wang, Zixin and Zhou, Yong and Shi, Yuanming and Letaief, Khaled B.},
  journal={IEEE Trans. Wirel. Commun.}, 
  title={Federated Fine-Tuning for Pre-Trained Foundation Models Over Wireless Networks}, 
  year={Apr. 2025},
  volume={24},
  number={4},
  pages={3450-3464}}

@INPROCEEDINGS{11044514,
  author={Qiu, Wenqi and Zhou, Yipeng and Wang, Jinzhi and Sheng, Quan Z. and Cui, Laizhong},
  booktitle={Proc. IEEE INFOCOM}, 
  title={{FLM}-{T}op{K}: Expediting Federated Large Language Model Tuning by Sparsifying Intervalized Gradients}, 
  year={London, UK, May 2025}
  }

@inproceedings{qin2023federated,
  title={Federated full-parameter tuning of billion-sized language models with communication cost under 18 kilobytes},
  author={Qin, Zhen and Chen, Daoyuan and Qian, Bingchen and Ding, Bolin and Li, Yaliang and Deng, Shuiguang},
  booktitle={Proc. Int. Conf. Mach. Learn. (ICML)},
  year={Honolulu, HI, Jul. 2024}
}

@inproceedings{shu2023zeroth,
  title={Zeroth-order optimization with trajectory-informed derivative estimation},
  author={Shu, Yao and Dai, Zhongxiang and Sng, Weicong and Verma, Arun and Jaillet, Patrick and Low, Bryan Kian Hsiang},
  booktitle={Proc. Int. Conf. Learn. Representations (ICLR)},
  year={Kigali, Rwanda, May 2023}
}

@inproceedings{malladi2023fine,
  title={Fine-tuning language models with just forward passes},
  author={Malladi, Sadhika and Gao, Tianyu and Nichani, Eshaan and Damian, Alex and Lee, Jason D and Chen, Danqi and Arora, Sanjeev},
  booktitle={Proc. Advances Neural Info. Process. Syst. (NeurIPS)},
  year={New Orleans, LA, Dec. 2023}
}

@inproceedings{chen2023deepzero,
  title={Deep{Z}ero: Scaling up zeroth-order optimization for deep model training},
  author={Chen, Aochuan and Zhang, Yimeng and Jia, Jinghan and Diffenderfer, James and Liu, Jiancheng and Parasyris, Konstantinos and Zhang, Yihua and Zhang, Zheng and Kailkhura, Bhavya and Liu, Sijia},
  booktitle={Proc. Int. Conf. Learn. Representations (ICLR)},
  year={Vienna, Austria, May 2024}
}

@inproceedings{chen2024enhancing,
  title={Enhancing zeroth-order fine-tuning for language models with low-rank structures},
  author={Chen, Yiming and Zhang, Yuan and Cao, Liyuan and Yuan, Kun and Wen, Zaiwen},
  booktitle={Proc. Int. Conf. Learn. Representations (ICLR)},
  year={Singapore, Apr. 2025}
}

@inproceedings{zhang2024revisiting,
  title={Revisiting zeroth-order optimization for memory-efficient {LLM} fine-tuning: A benchmark},
  author={Zhang, Yihua and Li, Pingzhi and Hong, Junyuan and Li, Jiaxiang and Zhang, Yimeng and Zheng, Wenqing and Chen, Pin-Yu and Lee, Jason D and Yin, Wotao and Hong, Mingyi and others},
  booktitle={Proc. Int. Conf. Mach. Learn. (ICML)},
  year={Vienna, Austria, Jul. 2024},
}

@inproceedings{ma2025revisiting,
  title={Revisiting zeroth-order optimization: Minimum-variance two-point estimators and directionally aligned perturbations},
  author={Ma, Shaocong and Huang, Heng},
  booktitle={Proc. Int. Conf. Learn. Representations (ICLR)},
  year={Singapore, Apr. 2025}
}

@ARTICLE{9917343,
  author={Fang, Wenzhi and Yu, Ziyi and Jiang, Yuning and Shi, Yuanming and Jones, Colin N. and Zhou, Yong},
  journal={IEEE Trans. Signal Process.}, 
  title={Communication-Efficient Stochastic Zeroth-Order Optimization for Federated Learning}, 
  year={Oct. 2022},
  volume={70},
  number={},
  pages={5058-5073}}

@inproceedings{qiu2023zeroth,
  title={Zeroth-order methods for nondifferentiable, nonconvex, and hierarchical federated optimization},
  author={Qiu, Yuyang and Shanbhag, Uday and Yousefian, Farzad},
  booktitle={Proc. Advances Neural Info. Process. Syst. (NeurIPS)},
  year={New Orleans, LA, Dec. 2023}
}

@inproceedings{chen2023fine,
  title={Fine-grained theoretical analysis of federated zeroth-order optimization},
  author={Chen, Jun and Chen, Hong and Gu, Bin and Deng, Hao},
  booktitle={Proc. Advances Neural Info. Process. Syst. (NeurIPS)},
  year={New Orleans, LA, Dec. 2023}
}

@inproceedings{ling2024convergence,
  title={On the convergence of zeroth-order federated tuning for large language models},
  author={Ling, Zhenqing and Chen, Daoyuan and Yao, Liuyi and Li, Yaliang and Shen, Ying},
  booktitle={Proc. ACM SIGKDD Conf. Knowl. Discovery Data Mining (SIGKDD)},
  year={Barcelona, Spain, Aug. 2024}
}

@inproceedings{li2024achieving,
  title={Achieving dimension-free communication in federated learning via zeroth-order optimization},
  author={Li, Zhe and Ying, Bicheng and Liu, Zidong and Dong, Chaosheng and Yang, Haibo},
  booktitle={Proc. Int. Conf. Mach. Learn. (ICML)},
  year={Vancouver, Canada, Jul. 2025}
}

@inproceedings{shu2024ferret,
  title={Ferret: Federated full-parameter tuning at scale for large language models},
  author={Shu, Yao and Hu, Wenyang and Ng, See-Kiong and Low, Bryan Kian Hsiang and Yu, Fei Richard},
  booktitle={Proc. Int. Conf. Mach. Learn. (ICML)},
  year={Vancouver, Canada, Jul. 2025}
}

@inproceedings{li2019convergence,
  title={On the Convergence of {F}ed{A}vg on Non-{IID} Data},
  author={Li, Xiang and Huang, Kaixuan and Yang, Wenhao and Wang, Shusen and Zhang, Zhihua},
  booktitle={Proc. Int. Conf. Learn. Representations (ICLR)},
  year={Apr. 2020}
}

@ARTICLE{10971879,
  author={Meng, Chuiyang and Tang, Ming and Setayesh, Mehdi and W.S. Wong, Vincent},
  journal={IEEE Trans. Mobile Comput.}, 
  title={Tackling Resource Allocation for Decentralized Federated Learning: A {GNN}-based Approach}, 
  year={Oct. 2025},
  volume={24},
  number={10},
  pages={9554-9569},}

@article{zhang2022opt,
  title={{OPT}: Open pre-trained transformer language models},
  author={Zhang, Susan and Roller, Stephen and Goyal, Naman and Artetxe, Mikel and Chen, Moya and Chen, Shuohui and Dewan, Christopher and Diab, Mona and Li, Xian and Lin, Xi Victoria and others},
  journal={arXiv preprint arXiv:2205.01068},
  year={May 2022}
}

@misc{AG-news,
  title = {{AG}'s corpus of news articles},
  author = {Antonio Gulli},
  howpublished = {\url{http://groups.di.unipi.it/~gulli/AG_corpus_of_news_articles.html}},
  note = {Accessed: 2025-07-30}
}

@inproceedings{socher2013recursive,
  title={Recursive deep models for semantic compositionality over a sentiment treebank},
  author={Socher, Richard and Perelygin, Alex and Wu, Jean and Chuang, Jason and Manning, Christopher D and Ng, Andrew Y and Potts, Christopher},
  booktitle={Proc. Conf. Empirical Methods Natural Lang. Process. (EMNLP)},
  year={Seattle, WA, Oct. 2013}
}

@inproceedings{bowman2015large,
  title={A large annotated corpus for learning natural language inference},
  author={Bowman, Samuel R and Angeli, Gabor and Potts, Christopher and Manning, Christopher D},
  Booktitle={Proc. Annu. Meeting Assoc. Comput. Linguistics (ACL)},
  year={Beijing, China, Jul. 2015}
}

@book{griva2008linear,
  title={Linear and Nonlinear Optimization 2nd edition},
  author={Griva, Igor and Nash, Stephen G and Sofer, Ariela},
  year={2008},
  publisher={Philadelphia, PA, USA: SIAM}
}

@article{nesterov2017random,
  title={Random gradient-free minimization of convex functions},
  author={Nesterov, Yurii and Spokoiny, Vladimir},
  journal={Foundations Comput. Math.},
  volume={17},
  number={2},
  pages={527--566},
  year={Apr. 2017}
}

@article{panchal2024thinking,
  title={Thinking forward: {M}emory-efficient federated finetuning of language models},
  author={Panchal, Kunjal and Parikh, Nisarg and Choudhary, Sunav and Zhang, Lijun and Brun, Yuriy and Guan, Hui},
  journal={Proc. Advances Neural Info. Process. Syst. (NeurIPS)},
  volume={},
  pages={},
  year={Vancouver, Canada, Dec. 2024}
}

@incollection{dinitz2006dinitz,
  title={Dinitz’algorithm: The original version and Even’s version},
  author={Dinitz, Yefim},
  booktitle={Theoretical Computer Science: Essays in Memory of Shimon Even},
  pages={218--240},
  year={2006},
  publisher={Berlin, Heidelberg, Germany: Springer}
}

@article{haimes1971bicriterion,
  title={On a bicriterion formulation of the problems of integrated system identification and system optimization},
  author={Haimes, Yacov},
  journal={IEEE Trans. Syst. Man Cybern.},
  number={3},
  pages={296--297},
  year={Jul. 1971},
  publisher={Institute of Electrical and Electronics Engineers (IEEE)}
}

\clearpage
\onecolumn
\appendix
\section*{Proof of Theorem 1}
\begin{proof}
According to the model update rule, the global model at the beginning of $(t+1)$-th fine-tuning round can be expressed as
\begin{align}
    \label{eq:global_model_t+1}
    \Bar{\mathbf{w}}^{t+1}
    = \left[\sum\limits_{n\in\mathcal{N}}\frac{a_{1,n}}{\sum\limits_{n'\in\mathcal{N}}a_{1,n'}}\left(\Bar{\mathbf{w}}_{1,n}^{t} -\eta^{t}\Tilde{\nabla} F_{n}\left(\Bar{\mathbf{w}}_{1,n}^{t};\xi_{n}\right)\right), \ldots, \sum\limits_{n\in\mathcal{N}}\frac{a_{M,n}}{\sum\limits_{n'\in\mathcal{N}}a_{M,n'}}\left(\Bar{\mathbf{w}}_{M,n}^{t} -\eta^{t}\Tilde{\nabla} F_{n}\left(\Bar{\mathbf{w}}_{M,n}^{t};\xi_{n}\right)\right)\right].
\end{align}
Based on Assumption 1, we expand $\mathbb{E}[f(\Bar{\mathbf{w}}^{t+1})]$ as
\begin{align}
    \label{eq:expand_1}
    \mathbb{E}\left[f(\Bar{\mathbf{w}}^{t+1})\right] \leq \mathbb{E}\left[f(\Bar{\mathbf{w}}^{t+1})\right] + \underbrace{\mathbb{E}\left[\left\langle\nabla f(\Bar{\mathbf{w}}^{t}), \Bar{\mathbf{w}}^{t+1} - \Bar{\mathbf{w}}^{t} \right\rangle\right]}_{T_{1}} + \frac{L}{2}\underbrace{\mathbb{E}\left[\left\Vert \Bar{\mathbf{w}}^{t+1} - \Bar{\mathbf{w}}^{t} \right\Vert^{2}\right]}_{T_{2}}.
\end{align}
We analyze $T_{1}$ first.
In particular, it satisfies
\begin{align}
    \label{eq:t1}
    &T_{1} \nonumber\\
    =&\: \mathbb{E}\left[\nabla f(\Bar{\mathbf{w}}^{t})^{\intercal}\right]\mathbb{E}\left[\Bar{\mathbf{w}}^{t+1} - \Bar{\mathbf{w}}^{t}\right] \nonumber\\
    =&\: \mathbb{E}\left[\nabla f(\Bar{\mathbf{w}}^{t})^{\intercal}\right]\mathbb{E}\left[-\eta^{t}\left[\sum\limits_{n\in\mathcal{N}}\frac{a_{1,n}}{\sum\limits_{n'\in\mathcal{N}}a_{1,n'}}\Tilde{\nabla} F_{n}\left(\Bar{\mathbf{w}}_{1,n}^{t};\xi_{n}\right), \ldots, \sum\limits_{n\in\mathcal{N}}\frac{a_{M,n}}{\sum\limits_{n'\in\mathcal{N}}a_{M,n'}}\Tilde{\nabla} F_{n}\left(\Bar{\mathbf{w}}_{M,n}^{t};\xi_{n}\right)\right]\right] \nonumber\\
    \overset{(\text{a})}{=}&\: -\eta^{t}\mathbb{E}\left[\nabla f(\Bar{\mathbf{w}}^{t})^{\intercal}\right]\mathbb{E}\left[\left[\sum\limits_{n\in\mathcal{N}}\frac{a_{1,n}}{\sum\limits_{n'\in\mathcal{N}}a_{1,n'}}\Tilde{\nabla} f_{n}\left(\Bar{\mathbf{w}}_{1,n}^{t}\right), \ldots, \sum\limits_{n\in\mathcal{N}}\frac{a_{M,n}}{\sum\limits_{n'\in\mathcal{N}}a_{M,n'}}\Tilde{\nabla} f_{n}\left(\Bar{\mathbf{w}}_{M,n}^{t}\right)\right]\right] \nonumber\\
    \overset{(\text{b})}{=}&\: -\frac{\eta^{t}}{2}\mathbb{E}\left[\left\Vert\nabla f(\Bar{\mathbf{w}}^{t})\right\Vert^{2}\right] - \frac{\eta^{t}}{2}\mathbb{E}\left[\left\Vert\left[\sum\limits_{n\in\mathcal{N}}\frac{a_{1,n}}{\sum\limits_{n'\in\mathcal{N}}a_{1,n'}}\Tilde{\nabla} f_{n}\left(\Bar{\mathbf{w}}_{1,n}^{t}\right), \ldots, \sum\limits_{n\in\mathcal{N}}\frac{a_{M,n}}{\sum\limits_{n'\in\mathcal{N}}a_{M,n'}}\Tilde{\nabla} f_{n}\left(\Bar{\mathbf{w}}_{M,n}^{t}\right)\right]\right\Vert^{2}\right]  \nonumber\\
    &+ \frac{\eta^{t}}{2}\underbrace{\mathbb{E}\left[\left\Vert\left[\nabla f(\Bar{\mathbf{w}}_{1}^{t}) - \sum\limits_{n\in\mathcal{N}}\frac{a_{1,n}}{\sum\limits_{n'\in\mathcal{N}}a_{1,n'}}\Tilde{\nabla} f_{n}\left(\Bar{\mathbf{w}}_{1,n}^{t}\right), \ldots, \nabla f(\Bar{\mathbf{w}}_{M}^{t}) - \sum\limits_{n\in\mathcal{N}}\frac{a_{M,n}}{\sum\limits_{n'\in\mathcal{N}}a_{M,n'}}\Tilde{\nabla} f_{n}\left(\Bar{\mathbf{w}}_{M,n}^{t}\right) \right] \right\Vert^{2}\right]}_{T_{3}},
\end{align}
where equality (a) results from Assumption 2. 
Equality (b) is obtained by using the fact that $\langle\mathbf{w}_{i}, \mathbf{w}_{j}\rangle = \frac{1}{2}(\Vert\mathbf{w}_{i}\Vert^{2} + \Vert\mathbf{w}_{j}\Vert^{2} - \Vert\mathbf{w}_{i}-\mathbf{w}_{j}\Vert^{2})$.
Then, we bound $T_{3}$. 
In particular, it satisfies
\begin{align}
    \label{eq:t3}
    &T_{3} \nonumber\\
    =&\: \mathbb{E}\Big[\Big\Vert\Big[
    \sum\limits_{n\in\mathcal{N}}\frac{a_{1,n}}{\sum\limits_{n'\in\mathcal{N}}a_{1,n'}}
    \Big(\nabla f(\Bar{\mathbf{w}}_{1}^{t}) \pm \nabla f_{n}(\Bar{\mathbf{w}}_{1}^{t}) - \Tilde{\nabla} f_{n}(\Bar{\mathbf{w}}_{1,n}^{t})\Big),
    \ldots, \nonumber\\
    &\hspace{3.7em}
    \sum\limits_{n\in\mathcal{N}}\frac{a_{M,n}}{\sum\limits_{n'\in\mathcal{N}}a_{M,n'}}
    \Big(\nabla f(\Bar{\mathbf{w}}_{M}^{t}) \pm \nabla f_{n}(\Bar{\mathbf{w}}_{M}^{t}) - \Tilde{\nabla} f_{n}(\Bar{\mathbf{w}}_{M,n}^{t})\Big)
    \Big]\Big\Vert^{2}\Big]\nonumber\\
    \overset{(\text{a})}{\leq}&\:
    2\mathbb{E}\left[\left\Vert\left[\sum\limits_{n\in\mathcal{N}}\frac{a_{1,n}}{\sum\limits_{n'\in\mathcal{N}}a_{1,n'}}\left(\nabla f(\Bar{\mathbf{w}}_{1}^{t}) - \nabla f_{n}(\Bar{\mathbf{w}}_{1}^{t})\right), \ldots, \sum\limits_{n\in\mathcal{N}}\frac{a_{M,n}}{\sum\limits_{n'\in\mathcal{N}}a_{M,n'}}\left(\nabla f(\Bar{\mathbf{w}}_{M}^{t}) - \nabla f_{n}(\Bar{\mathbf{w}}_{M}^{t}) \right) \right] \right\Vert^{2}\right] \nonumber\\
    &+ 2\mathbb{E}\left[\left\Vert\left[\sum\limits_{n\in\mathcal{N}}\frac{a_{1,n}}{\sum\limits_{n'\in\mathcal{N}}a_{1,n'}}\left(\nabla f_{n}(\Bar{\mathbf{w}}_{1}^{t}) - \Tilde{\nabla} f_{n}\left(\Bar{\mathbf{w}}_{1,n}^{t}\right)\right), \ldots, \sum\limits_{n\in\mathcal{N}}\frac{a_{M,n}}{\sum\limits_{n'\in\mathcal{N}}a_{M,n'}}\left(\nabla f_{n}(\Bar{\mathbf{w}}_{M}^{t}) - \Tilde{\nabla} f_{n}\left(\Bar{\mathbf{w}}_{M,n}^{t}\right)\right) \right] \right\Vert^{2}\right] \nonumber\\
    \overset{(\text{b})}{\leq}&\:
    2N\sum\limits_{n\in\mathcal{N}}\mathbb{E}\left[\left\Vert\left[\frac{a_{1,n}}{\sum\limits_{n'\in\mathcal{N}}a_{1,n'}}\left(\nabla f(\Bar{\mathbf{w}}_{1}^{t}) - \nabla f_{n}(\Bar{\mathbf{w}}_{1}^{t})\right), \ldots, \frac{a_{M,n}}{\sum\limits_{n'\in\mathcal{N}}a_{M,n'}}\left(\nabla f(\Bar{\mathbf{w}}_{M}^{t}) - \nabla f_{n}(\Bar{\mathbf{w}}_{M}^{t}) \right) \right] \right\Vert^{2}\right] \nonumber\\
    &+ 2N\sum\limits_{n\in\mathcal{N}}\mathbb{E}\left[\left\Vert\left[\frac{a_{1,n}}{\sum\limits_{n'\in\mathcal{N}}a_{1,n'}}\left(\nabla f_{n}(\Bar{\mathbf{w}}_{1}^{t}) - \Tilde{\nabla} f_{n}\left(\Bar{\mathbf{w}}_{1,n}^{t}\right)\right), \ldots, \frac{a_{M,n}}{\sum\limits_{n'\in\mathcal{N}}a_{M,n'}}\left(\nabla f_{n}(\Bar{\mathbf{w}}_{M}^{t}) - \Tilde{\nabla} f_{n}\left(\Bar{\mathbf{w}}_{M,n}^{t}\right)\right) \right] \right\Vert^{2}\right] \nonumber\\
    \overset{(\text{c})}{\leq}&\: 
    2N\sum\limits_{n\in\mathcal{N}}\max\limits_{m\in\mathcal{M}}\left\{\left(\frac{a_{m,n}}{\sum\limits_{n'\in\mathcal{N}}a_{m,n'}}\right)^{2}\right\}\mathbb{E}\left[\left\Vert\left(\nabla f(\Bar{\mathbf{w}}^{t}) - \nabla f_{n}(\Bar{\mathbf{w}}^{t})\right) \right\Vert^{2}\right] \nonumber\\
    &+ 2N\sum\limits_{n\in\mathcal{N}}\max\limits_{m\in\mathcal{M}}\left\{\left(\frac{a_{m,n}}{\sum\limits_{n'\in\mathcal{N}}a_{m,n'}}\right)^{2}\right\}\mathbb{E}\left[\left\Vert\left(\nabla f_{n}(\Bar{\mathbf{w}}^{t}) - \Tilde{\nabla} f_{n}(\Bar{\mathbf{w}}^{t})\right) \right\Vert^{2}\right] \nonumber\\
    \overset{(\text{d})}{\leq}&\: 2N\left(\sigma_{G}^{2} + \frac{1}{4}\mu^{2}L^{2}(d+3)^{3}\right)\sum\limits_{n\in\mathcal{N}}\max\limits_{m\in\mathcal{M}}\left\{\left(\frac{a_{m,n}}{\sum\limits_{n'\in\mathcal{N}}a_{m,n'}}\right)^{2}\right\},
\end{align}
where inequalities (a) and (b) result from Jensen's inequality.
Inequality (d) is obtained from Assumption 3 and Lemma 1.
Therefore, by combining inequality (\ref{eq:t3}) and inequality (\ref{eq:t1}), we have
\begin{align}
    \label{eq:t1_final}
    T_{1}
    \leq&\: -\frac{\eta^{t}}{2}\mathbb{E}\left[\left\Vert\nabla f(\Bar{\mathbf{w}}^{t})\right\Vert^{2}\right] - \frac{\eta^{t}}{2}\mathbb{E}\left[\left\Vert\left[\sum\limits_{n\in\mathcal{N}}\frac{a_{1,n}}{\sum\limits_{n'\in\mathcal{N}}a_{1,n'}}\Tilde{\nabla} f_{n}\left(\Bar{\mathbf{w}}_{1,n}^{t}\right), \ldots, \sum\limits_{n\in\mathcal{N}}\frac{a_{M,n}}{\sum\limits_{n'\in\mathcal{N}}a_{M,n'}}\Tilde{\nabla} f_{n}\left(\Bar{\mathbf{w}}_{M,n}^{t}\right)\right]\right\Vert^{2}\right] \nonumber\\
    &+ \eta^{t}N\left(\sigma_{G}^{2} + \frac{1}{4}\mu^{2}L^{2}(d+3)^{3}\right)\sum\limits_{n\in\mathcal{N}}\max\limits_{m\in\mathcal{M}}\left\{\left(\frac{a_{m,n}}{\sum\limits_{n'\in\mathcal{N}}a_{m,n'}}\right)^{2}\right\} \nonumber\\
    \overset{(\text{a})}{\leq}&\: -\frac{\eta^{t}}{2}\mathbb{E}\left[\left\Vert\nabla f(\Bar{\mathbf{w}}^{t})\right\Vert^{2}\right] + \eta^{t}N\left(\sigma_{G}^{2} + \frac{1}{4}\mu^{2}L^{2}(d+3)^{3}\right)\sum\limits_{n\in\mathcal{N}}\max\limits_{m\in\mathcal{M}}\left\{\left(\frac{a_{m,n}}{\sum\limits_{n'\in\mathcal{N}}a_{m,n'}}\right)^{2}\right\},
\end{align}
where inequality (a) is because $\mathbb{E}\left[\left\Vert\left[\sum\limits_{n\in\mathcal{N}}\frac{a_{1,n}}{\sum\limits_{n'\in\mathcal{N}}a_{1,n'}}\Tilde{\nabla} f_{n}\left(\Bar{\mathbf{w}}_{1,n}^{t}\right), \ldots, \sum\limits_{n\in\mathcal{N}}\frac{a_{M,n}}{\sum\limits_{n'\in\mathcal{N}}a_{M,n'}}\Tilde{\nabla} f_{n}\left(\Bar{\mathbf{w}}_{M,n}^{t}\right)\right]\right\Vert^{2}\right] \geq 0$.
Then, we bound $T_{2}$.
In particular, it satisfies
\begin{align}
    \label{eq:t2}
    &T_{2} \nonumber\\
    =&\: \mathbb{E}\left[\left\Vert-\eta^{t}\left[\sum\limits_{n\in\mathcal{N}}\frac{a_{1,n}}{\sum\limits_{n'\in\mathcal{N}}a_{1,n'}}\Tilde{\nabla} F_{n}\left(\Bar{\mathbf{w}}_{1,n}^{t};\xi_{n}\right), \ldots, \sum\limits_{n\in\mathcal{N}}\frac{a_{M,n}}{\sum\limits_{n'\in\mathcal{N}}a_{M,n'}}\Tilde{\nabla} F_{n}\left(\Bar{\mathbf{w}}_{M,n}^{t};\xi_{n}\right)\right]\right\Vert^{2}\right] \nonumber\\
    \overset{(\text{a})}{\leq}&\:\left(\eta^{t}\right)^{2}N\sum\limits_{n\in\mathcal{N}}\mathbb{E}\left[\left\Vert\left[\frac{a_{1,n}}{\sum\limits_{n'\in\mathcal{N}}a_{1,n'}}\Tilde{\nabla} F_{n}\left(\Bar{\mathbf{w}}_{1,n}^{t};\xi_{n}\right), \ldots, \frac{a_{M,n}}{\sum\limits_{n'\in\mathcal{N}}a_{M,n'}}\Tilde{\nabla} F_{n}\left(\Bar{\mathbf{w}}_{M,n}^{t};\xi_{n}\right)\right]\right\Vert^{2}\right] \nonumber\\
    \overset{(\text{b})}{\leq}&\: \left(\eta^{t}\right)^{2}N\sum\limits_{n\in\mathcal{N}}\max\limits_{m\in\mathcal{M}}\left\{\left(\frac{a_{m,n}}{\sum\limits_{n'\in\mathcal{N}}a_{m,n'}}\right)^{2}\right\}\mathbb{E}\left[\left\Vert\Tilde{\nabla} F_{n}\left(\Bar{\mathbf{w}}^{t};\xi_{n}\right)\right\Vert^{2}\right] \nonumber\\
    \overset{(\text{c})}{\leq}&\: \left(\eta^{t}\right)^{2}dN\sum\limits_{n\in\mathcal{N}}\max\limits_{m\in\mathcal{M}}\left\{\left(\frac{a_{m,n}}{\sum\limits_{n'\in\mathcal{N}}a_{m,n'}}\right)^{2}\right\}\mathbb{E}\left[\left\Vert\nabla F_{n}\left(\Bar{\mathbf{w}}^{t};\xi_{n}\right)\right\Vert^{2}\right] \nonumber\\
    =&\: \left(\eta^{t}\right)^{2}dN\sum\limits_{n\in\mathcal{N}}\max\limits_{m\in\mathcal{M}}\left\{\left(\frac{a_{m,n}}{\sum\limits_{n'\in\mathcal{N}}a_{m,n'}}\right)^{2}\right\}\mathbb{E}\left[\left\Vert\nabla F_{n}\left(\Bar{\mathbf{w}}^{t};\xi_{n}\right) \pm \nabla f_{n}\left(\Bar{\mathbf{w}}^{t}\right) \pm \nabla f\left(\Bar{\mathbf{w}}^{t}\right)\right\Vert^{2}\right] \nonumber\\
    \overset{(\text{d})}{\leq}&\: 3\left(\eta^{t}\right)^{2}dN\sum\limits_{n\in\mathcal{N}}\max\limits_{m\in\mathcal{M}}\left\{\left(\frac{a_{m,n}}{\sum\limits_{n'\in\mathcal{N}}a_{m,n'}}\right)^{2}\right\}\Big(\mathbb{E}\Big[\left\Vert\nabla F_{n}\left(\Bar{\mathbf{w}}^{t};\xi_{n}\right) - \nabla f_{n}\left(\Bar{\mathbf{w}}^{t}\right)\right\Vert^{2}\Big] \nonumber\\
    &\mathbb{E}\Big[\left\Vert\nabla f_{n}\left(\Bar{\mathbf{w}}^{t}\right) - \nabla f\left(\Bar{\mathbf{w}}^{t}\right)\right\Vert^{2}\Big] + \mathbb{E}\Big[\left\Vert\nabla f\left(\Bar{\mathbf{w}}^{t}\right)\right\Vert^{2}\Big]\Big) \nonumber\\
    \overset{(\text{e})}{\leq}&\:3\left(\eta^{t}\right)^{2}dN\sum\limits_{n\in\mathcal{N}}\max\limits_{m\in\mathcal{M}}\left\{\left(\frac{a_{m,n}}{\sum\limits_{n'\in\mathcal{N}}a_{m,n'}}\right)^{2}\right\}\left(\mathbb{E}\Big[\left\Vert\nabla f\left(\Bar{\mathbf{w}}^{t}\right)\right\Vert^{2}\Big] + \sigma^{2} + \sigma_{G}^{2}\right),
\end{align}
where inequalities (a) and (d) result from Jensen's inequality.
Inequality (b) is obtained by using Lemma 2.
Inequality (c) is derived by using Lemma 3.
Inequality (e) results from Assumptions 2 and 3.
Then, we combine inequalities (\ref{eq:expand_1}), (\ref{eq:t1_final}), and (\ref{eq:t2}) as follows:
\begin{align}
    \label{eq:expand_2}
    &\mathbb{E}\left[f(\Bar{\mathbf{w}}^{t+1})\right] \nonumber\\
    \leq&\: \mathbb{E}\left[f(\Bar{\mathbf{w}}^{t})\right] -\frac{\eta^{t}}{2}\mathbb{E}\left[\left\Vert\nabla f(\Bar{\mathbf{w}}^{t})\right\Vert^{2}\right] + \eta^{t}N\left(\sigma_{G}^{2} + \frac{1}{4}\mu^{2}L^{2}(d+3)^{3}\right)\sum\limits_{n\in\mathcal{N}}\max\limits_{m\in\mathcal{M}}\left\{\left(\frac{a_{m,n}}{\sum\limits_{n'\in\mathcal{N}}a_{m,n'}}\right)^{2}\right\} \nonumber\\
    &+ \frac{L}{2}\left(3\left(\eta^{t}\right)^{2}dN\sum\limits_{n\in\mathcal{N}}\max\limits_{m\in\mathcal{M}}\left\{\left(\frac{a_{m,n}}{\sum\limits_{n'\in\mathcal{N}}a_{m,n'}}\right)^{2}\right\}\left(\mathbb{E}\Big[\left\Vert\nabla f\left(\Bar{\mathbf{w}}^{t}\right)\right\Vert^{2}\Big] + \sigma^{2} + \sigma_{G}^{2}\right)\right).
\end{align}
By rearranging inequality (\ref{eq:expand_2}), we have
\begin{align}
    \label{eq:expand_3}
    &\left(\frac{\eta^{t}}{2} - \frac{3L\left(\eta^{t}\right)^{2}dN}{2}\sum\limits_{n\in\mathcal{N}}\max\limits_{m\in\mathcal{M}}\left\{\left(\frac{a_{m,n}}{\sum\limits_{n'\in\mathcal{N}}a_{m,n'}}\right)^{2}\right\}\right)\mathbb{E}\Big[\left\Vert\nabla f\left(\Bar{\mathbf{w}}^{t}\right)\right\Vert^{2}\Big] \nonumber\\
    \leq&\: \mathbb{E}\left[f(\Bar{\mathbf{w}}^{t})\right] - \mathbb{E}\left[f(\Bar{\mathbf{w}}^{t+1})\right] + \eta^{t}N\left(\sigma_{G}^{2} + \frac{1}{4}\mu^{2}L^{2}(d+3)^{3}\right)\sum\limits_{n\in\mathcal{N}}\max\limits_{m\in\mathcal{M}}\left\{\left(\frac{a_{m,n}}{\sum\limits_{n'\in\mathcal{N}}a_{m,n'}}\right)^{2}\right\} \nonumber\\
    &+ \frac{L}{2}\left(3\left(\eta^{t}\right)^{2}dN\left(\sigma^{2} + \sigma_{G}^{2}\right)\sum\limits_{n\in\mathcal{N}}\max\limits_{m\in\mathcal{M}}\left\{\left(\frac{a_{m,n}}{\sum\limits_{n'\in\mathcal{N}}a_{m,n'}}\right)^{2}\right\}\right).
\end{align}
We define $\Omega(\mathbf{A}) = \left(\frac{\eta^{t}}{2} - \frac{3L\left(\eta^{t}\right)^{2}dN}{2}\sum\limits_{n\in\mathcal{N}}\max\limits_{m\in\mathcal{M}}\left\{\left(\frac{a_{m,n}}{\sum\limits_{n'\in\mathcal{N}}a_{m,n'}}\right)^{2}\right\}\right)$.
To guarantee $\Omega(\mathbf{A}) \geq 0$, the learning rate in the arbitrary $t$-th fine-tuning round must be small enough such that 
\begin{align}
    \label{eq:lr}
    \eta^{t} = \eta \leq \min\limits_{\mathbf{A}}\Bigg\{\frac{1}{3LdN\sum\limits_{n\in\mathcal{N}}\max\limits_{m\in\mathcal{M}}\Big\{\big(\frac{a_{m,n}}{\sum\limits_{n'\in\mathcal{N}}a_{m,n'}}\big)^{2}\Big\}}\Bigg\} = \frac{1}{3LdN^{2}}.
\end{align}
Hence, we have
\begin{align}
    \label{eq:expand_4}
    \mathbb{E}\Big[\left\Vert\nabla f\left(\Bar{\mathbf{w}}^{t}\right)\right\Vert^{2}\Big] 
    \leq \frac{1}{\Omega(\mathbf{A})}\left(\mathbb{E}\left[f(\Bar{\mathbf{w}}^{t})\right] - \mathbb{E}\left[f(\Bar{\mathbf{w}}^{t+1})\right]\right) + \frac{\Theta_{1}}{\Omega(\mathbf{A})}\sum\limits_{n\in\mathcal{N}}\max\limits_{m\in\mathcal{M}}\left\{\left(\frac{a_{m,n}}{\sum\limits_{n'\in\mathcal{N}}a_{m,n'}}\right)^{2}\right\},
\end{align}
where $\Theta_{1} = \frac{1}{4}\mu^{2}L^{2}\eta(d+3)^{3}N + \left(\eta N + \frac{3L\eta^{2}dN}{2}\right)\sigma_{G}^{2} + \frac{3L\eta^{2}dN}{2}\sigma^{2}$.
To determine the convergence rate, we sum up both sides of inequality (\ref{eq:expand_4}) for all $T$ fine-tuning rounds and multiply both sides by $\frac{1}{T}$.
We have
\begin{align}
    \label{eq:std_results}
    \frac{1}{T}\sum\limits_{t=1}^{T}\mathbb{E}\Big[\left\Vert\nabla f\left(\Bar{\mathbf{w}}^{t}\right)\right\Vert^{2}\Big] 
    \leq&\: \frac{1}{\Omega(\mathbf{A})T}\sum\limits_{t=1}^{T}\left(f(\Bar{\mathbf{w}}^{t}) - f(\Bar{\mathbf{w}}^{t+1})\right) + \frac{\Theta_{1}}{\Omega(\mathbf{A})}\sum\limits_{n\in\mathcal{N}}\max\limits_{m\in\mathcal{M}}\left\{\left(\frac{a_{m,n}}{\sum\limits_{n'\in\mathcal{N}}a_{m,n'}}\right)^{2}\right\} \nonumber\\
    \overset{(\text{a})}{\leq}&\: \frac{1}{\Omega(\mathbf{A})T}\left(f(\Bar{\mathbf{w}}^{1}) - f(\mathbf{w}^{\star})\right) + \frac{\Theta_{1}}{\Omega(\mathbf{A})}\sum\limits_{n\in\mathcal{N}}\max\limits_{m\in\mathcal{M}}\left\{\left(\frac{a_{m,n}}{\sum\limits_{n'\in\mathcal{N}}a_{m,n'}}\right)^{2}\right\},
\end{align}
where inequality (a) is obtained by using $f(\bar{\mathbf{w}}^{T+1}) \geq f(\mathbf{w}^{\star})$.
This completes the proof of Theorem 1.
\end{proof}

\section*{Proof of Theorem 2}
\begin{proof}
Based on second-order Taylor expansion, we expand $\mathbb{E}[f(\Bar{\mathbf{w}}^{t+1})]$ as
\begin{align}
    \label{eq:expand_5}
    &\mathbb{E}\left[f(\Bar{\mathbf{w}}^{t+1})\right] \nonumber\\
    \approx&\: \mathbb{E}\left[f(\Bar{\mathbf{w}}^{t})\right] + \mathbb{E}\left[\left\langle\nabla f(\Bar{\mathbf{w}}^{t}), \Bar{\mathbf{w}}^{t+1} - \Bar{\mathbf{w}}^{t} \right\rangle\right] + \frac{1}{2}\mathbb{E}\left[(\Bar{\mathbf{w}}^{t+1} - \Bar{\mathbf{w}}^{t})^{\intercal}\nabla^{2}f(\Bar{\mathbf{w}}^{t})(\Bar{\mathbf{w}}^{t+1} - \Bar{\mathbf{w}}^{t})\right] \nonumber\\
    \overset{(\text{a})}{\leq}&\: \mathbb{E}\left[f(\Bar{\mathbf{w}}^{t})\right] -\frac{\eta^{t}}{2}\mathbb{E}\left[\left\Vert\nabla f(\Bar{\mathbf{w}}^{t})\right\Vert^{2}\right] + \eta^{t}N\left(\sigma_{G}^{2} + \frac{1}{4}\mu^{2}L^{2}(d+3)^{3}\right)\sum\limits_{n\in\mathcal{N}}\max\limits_{m\in\mathcal{M}}\left\{\left(\frac{a_{m,n}}{\sum\limits_{n'\in\mathcal{N}}a_{m,n'}}\right)^{2}\right\} \nonumber\\
    &+ \frac{1}{2}\mathbb{E}\left[(\Bar{\mathbf{w}}^{t+1} - \Bar{\mathbf{w}}^{t})^{\intercal}\nabla^{2}f(\Bar{\mathbf{w}}^{t})(\Bar{\mathbf{w}}^{t+1} - \Bar{\mathbf{w}}^{t})\right] \nonumber\\
    \overset{(\text{b})}{\leq}&\: \mathbb{E}\left[f(\Bar{\mathbf{w}}^{t})\right] -\frac{\eta^{t}}{2}\mathbb{E}\left[\left\Vert\nabla f(\Bar{\mathbf{w}}^{t})\right\Vert^{2}\right] + \eta^{t}N\left(\sigma_{G}^{2} + \frac{1}{4}\mu^{2}L^{2}(d+3)^{3}\right)\sum\limits_{n\in\mathcal{N}}\max\limits_{m\in\mathcal{M}}\left\{\left(\frac{a_{m,n}}{\sum\limits_{n'\in\mathcal{N}}a_{m,n'}}\right)^{2}\right\} \nonumber\\
    &+ \frac{1}{2}\underbrace{\mathbb{E}\left[\left\langle\mathbf{H}(\Bar{\mathbf{w}}^{t}), (\Bar{\mathbf{w}}^{t+1} - \Bar{\mathbf{w}}^{t})(\Bar{\mathbf{w}}^{t+1} - \Bar{\mathbf{w}}^{t})^{\intercal}\right\rangle\right]}_{T_{1}}, \nonumber\\
    =&\: \mathbb{E}\left[f(\Bar{\mathbf{w}}^{t})\right]-\frac{\eta^{t}}{2}\mathbb{E}\left[\left\Vert\nabla f(\Bar{\mathbf{w}}^{t})\right\Vert^{2}\right] + \eta^{t}N\left(\sigma_{G}^{2} + \frac{1}{4}\mu^{2}L^{2}(d+3)^{3}\right)\sum\limits_{n\in\mathcal{N}}\max\limits_{m\in\mathcal{M}}\left\{\left(\frac{a_{m,n}}{\sum\limits_{n'\in\mathcal{N}}a_{m,n'}}\right)^{2}\right\} \nonumber\\
    &+ \frac{1}{2}\mathbb{E}\left[\mathbf{H}(\Bar{\mathbf{w}}^{t})\right]\underbrace{\mathbb{E}\left[(\Bar{\mathbf{w}}^{t+1} - \Bar{\mathbf{w}}^{t})(\Bar{\mathbf{w}}^{t+1} - \Bar{\mathbf{w}}^{t})^{\intercal}\right]}_{T_{2}},
\end{align}
where inequality (a) results from inequality (\ref{eq:t3}) in the proof of Theorem 1.
Inequality (b) is obtained by using Assumption 4.
Then, to bound $T_{1}$, we analyze $T_{2}$.
We define $\mathbf{\Sigma}(\mathbf{w}) = \mathbb{E}\left[\nabla F(\mathbf{w}; \bm{\xi})\left(\nabla F(\mathbf{w}; \bm{\xi})\right)^{\intercal}\right] - \nabla f(\mathbf{w})\left(\nabla f(\mathbf{w})\right)^{\intercal}$.
In particular, $T_{2}$ satisfies
\begin{align}
    \label{eq:new_t1}
    &T_{2} \nonumber\\
    \overset{(\text{a})}{\leq}&\: \left(1 + \frac{d-2}{d+2}\right)(\eta^{t})^{2}N\sum\limits_{n\in\mathcal{N}}\max\limits_{m\in\mathcal{M}}\left\{\left(\frac{a_{m,n}}{\sum\limits_{n'\in\mathcal{N}}a_{m,n'}}\right)^{2}\right\}\left(\nabla f(\Bar{\mathbf{w}}^{t})\left(\nabla f(\Bar{\mathbf{w}}^{t})\right)^{\intercal} + \frac{1}{N}\mathbf{\Sigma}(\Bar{\mathbf{w}}^{t})\right) \nonumber\\
    &+ \frac{d}{d+2}(\eta^{t})^{2}N\sum\limits_{n\in\mathcal{N}}\max\limits_{m\in\mathcal{M}}\left\{\left(\frac{a_{m,n}}{\sum\limits_{n'\in\mathcal{N}}a_{m,n'}}\right)^{2}\right\}\left( \left\Vert\nabla f(\Bar{\mathbf{w}}^{t})\right\Vert^{2} + \frac{1}{N}\Tr\left(\mathbf{\Sigma}(\Bar{\mathbf{w}}^{t})\right)\right)\mathbf{I}_{d\times d},
\end{align}
where inequality (a) results from Jensen's inequality and Lemma 3.
By combining inequalities (\ref{eq:expand_5}) and (\ref{eq:new_t1}), we have
\begin{align}
    \label{eq:expand_6}
    &\mathbb{E}\left[f(\Bar{\mathbf{w}}^{t+1})\right] \nonumber\\
    \leq&\: \mathbb{E}\left[f(\Bar{\mathbf{w}}^{t})\right] -\frac{\eta^{t}}{2}\mathbb{E}\left[\left\Vert\nabla f(\Bar{\mathbf{w}}^{t})\right\Vert^{2}\right] + \eta^{t}N\left(\sigma_{G}^{2} + \frac{1}{4}\mu^{2}L^{2}(d+3)^{3}\right)\sum\limits_{n\in\mathcal{N}}\max\limits_{m\in\mathcal{M}}\left\{\left(\frac{a_{m,n}}{\sum\limits_{n'\in\mathcal{N}}a_{m,n'}}\right)^{2}\right\} \nonumber\\
    &+ \frac{1}{2}\left(1 + \frac{d-2}{d+2}\right)(\eta^{t})^{2}N\sum\limits_{n\in\mathcal{N}}\max\limits_{m\in\mathcal{M}}\left\{\left(\frac{a_{m,n}}{\sum\limits_{n'\in\mathcal{N}}a_{m,n'}}\right)^{2}\right\}\left(\nabla f(\Bar{\mathbf{w}}^{t})^{\intercal}\mathbf{H}(\Bar{\mathbf{w}}^{t})\nabla f(\Bar{\mathbf{w}}^{t}) + \left\langle\frac{1}{N}\mathbf{\Sigma}(\Bar{\mathbf{w}}^{t}), \mathbf{H}(\Bar{\mathbf{w}}^{t})\right\rangle\right) \nonumber\\
    &+ \frac{d}{2(d+2)}(\eta^{t})^{2}N\sum\limits_{n\in\mathcal{N}}\max\limits_{m\in\mathcal{M}}\left\{\left(\frac{a_{m,n}}{\sum\limits_{n'\in\mathcal{N}}a_{m,n'}}\right)^{2}\right\}\left( \left\Vert\nabla f(\Bar{\mathbf{w}}^{t})\right\Vert^{2} + \frac{1}{N}\Tr\left(\mathbf{\Sigma}(\Bar{\mathbf{w}}^{t})\right)\right)\Tr\left(\mathbf{H}(\Bar{\mathbf{w}}^{t})\right) \nonumber\\
    \overset{(\text{a})}{\leq}&\: \mathbb{E}\left[f(\Bar{\mathbf{w}}^{t})\right] -\frac{\eta^{t}}{2}\mathbb{E}\left[\left\Vert\nabla f(\Bar{\mathbf{w}}^{t})\right\Vert^{2}\right] + \eta^{t}N\left(\sigma_{G}^{2} + \frac{1}{4}\mu^{2}L^{2}(d+3)^{3}\right)\sum\limits_{n\in\mathcal{N}}\max\limits_{m\in\mathcal{M}}\left\{\left(\frac{a_{m,n}}{\sum\limits_{n'\in\mathcal{N}}a_{m,n'}}\right)^{2}\right\} \nonumber\\
    &+ \frac{1}{2}\left(1 + \frac{d-2}{d+2}\right)(\eta^{t})^{2}NL\sum\limits_{n\in\mathcal{N}}\max\limits_{m\in\mathcal{M}}\left\{\left(\frac{a_{m,n}}{\sum\limits_{n'\in\mathcal{N}}a_{m,n'}}\right)^{2}\right\}\left(\left\Vert\nabla f(\Bar{\mathbf{w}}^{t})\right\Vert^{2} + \frac{1}{N}\Tr\left(\mathbf{\Sigma}(\Bar{\mathbf{w}}^{t})\right)\right) \nonumber\\
    &+ \frac{d}{2(d+2)}(\eta^{t})^{2}NL\kappa\sum\limits_{n\in\mathcal{N}}\max\limits_{m\in\mathcal{M}}\left\{\left(\frac{a_{m,n}}{\sum\limits_{n'\in\mathcal{N}}a_{m,n'}}\right)^{2}\right\}\left(\left\Vert\nabla f(\Bar{\mathbf{w}}^{t})\right\Vert^{2} + \frac{1}{N}\Tr\left(\mathbf{\Sigma}(\Bar{\mathbf{w}}^{t})\right)\right) \nonumber\\
    \overset{(\text{b})}{\leq}&\: \mathbb{E}\left[f(\Bar{\mathbf{w}}^{t})\right] -\frac{\eta^{t}}{2}\mathbb{E}\left[\left\Vert\nabla f(\Bar{\mathbf{w}}^{t})\right\Vert^{2}\right] + \eta^{t}N\left(\sigma_{G}^{2} + \frac{1}{4}\mu^{2}L^{2}(d+3)^{3}\right)\sum\limits_{n\in\mathcal{N}}\max\limits_{m\in\mathcal{M}}\left\{\left(\frac{a_{m,n}}{\sum\limits_{n'\in\mathcal{N}}a_{m,n'}}\right)^{2}\right\} \nonumber\\
    &+ \frac{d(\eta^{t})^{2}NL(\kappa+2)}{2(d+2)}\sum\limits_{n\in\mathcal{N}}\max\limits_{m\in\mathcal{M}}\left\{\left(\frac{a_{m,n}}{\sum\limits_{n'\in\mathcal{N}}a_{m,n'}}\right)^{2}\right\}\left(\left\Vert\nabla f(\Bar{\mathbf{w}}^{t})\right\Vert^{2} + \frac{\sigma^{2}}{N}\right),
\end{align}
where inequality (a) results from Assumption 4.
Inequality (b) is obtained by using the fact that $\Tr\left(\mathbf{\Sigma}(\mathbf{w})\right) = \mathbb{E}\big[\Vert\nabla F(\mathbf{w}; \bm{\xi}) - \nabla f(\mathbf{w}) \Vert^{2}\big]$ and Assumption 2.
By rearranging inequality (\ref{eq:expand_6}), we have
\begin{align}
    \label{eq:expand_7}
    &\left(\frac{\eta^{t}}{2} - \frac{d(\eta^{t})^{2}NL(\kappa+2)}{2(d+2)}\sum\limits_{n\in\mathcal{N}}\max\limits_{m\in\mathcal{M}}\left\{\left(\frac{a_{m,n}}{\sum\limits_{n'\in\mathcal{N}}a_{m,n'}}\right)^{2}\right\}\right)\left\Vert\nabla f(\Bar{\mathbf{w}}^{t})\right\Vert^{2} \nonumber\\
    \leq&\: \mathbb{E}\left[f(\Bar{\mathbf{w}}^{t})\right] - \mathbb{E}\left[f(\Bar{\mathbf{w}}^{t+1})\right] + \eta^{t}N\left(\sigma_{G}^{2} + \frac{1}{4}\mu^{2}L^{2}(d+3)^{3}\right)\sum\limits_{n\in\mathcal{N}}\max\limits_{m\in\mathcal{M}}\left\{\left(\frac{a_{m,n}}{\sum\limits_{n'\in\mathcal{N}}a_{m,n'}}\right)^{2}\right\} \nonumber\\
    &+ \frac{d(\eta^{t})^{2}L(\kappa+2)\sigma^{2}}{2(d+2)}\sum\limits_{n\in\mathcal{N}}\max\limits_{m\in\mathcal{M}}\left\{\left(\frac{a_{m,n}}{\sum\limits_{n'\in\mathcal{N}}a_{m,n'}}\right)^{2}\right\}.
\end{align}
We define $\Phi(\mathbf{A}) = \left(\frac{\eta^{t}}{2} - \frac{d(\eta^{t})^{2}NL(\kappa+2)}{2(d+2)}\sum\limits_{n\in\mathcal{N}}\max\limits_{m\in\mathcal{M}}\left\{\left(\frac{a_{m,n}}{\sum\limits_{n'\in\mathcal{N}}a_{m,n'}}\right)^{2}\right\}\right)$.
To guarantee $\Phi(\mathbf{A}) \geq 0$, the learning rate in the arbitrary $t$-th fine-tuning round must be small enough such that
\begin{align}
    \label{eq:lr_dimfree}
    \eta^{t} = \eta \leq \min\limits_{\mathbf{A}}\Bigg\{\frac{d+2}{LdN(\kappa+2)\sum\limits_{n\in\mathcal{N}}\max\limits_{m\in\mathcal{M}}\Big\{\big(\frac{a_{m,n}}{\sum\limits_{n'\in\mathcal{N}}a_{m,n'}}\big)^{2}\Big\}}\Bigg\} = \frac{d+2}{LdN^{2}(\kappa+2)}.
\end{align}
Hence, we have
\begin{align}
    \label{eq:expand_8}
    \mathbb{E}\Big[\left\Vert\nabla f\left(\Bar{\mathbf{w}}^{t}\right)\right\Vert^{2}\Big] 
    \leq\: \frac{1}{\Phi(\mathbf{A})}\left(\mathbb{E}\left[f(\Bar{\mathbf{w}}^{t})\right] - \mathbb{E}\left[f(\Bar{\mathbf{w}}^{t+1})\right]\right) + \frac{\Theta_{2}}{\Phi(\mathbf{A})}\sum\limits_{n\in\mathcal{N}}\max\limits_{m\in\mathcal{M}}\left\{\left(\frac{a_{m,n}}{\sum\limits_{n'\in\mathcal{N}}a_{m,n'}}\right)^{2}\right\},
\end{align}
where $\Theta_{2} = \eta^{t}N\left(\sigma_{G}^{2} + \frac{1}{4}\mu^{2}L^{2}(d+3)^{3}\right) + \frac{d(\eta^{t})^{2}L(\kappa+2)\sigma^{2}}{2(d+2)}$.
To determine the convergence rate, we sum up both sides of inequality (\ref{eq:expand_8}) for all $T$ fine-tuning rounds and multiply both sides by $\frac{1}{T}$.
We have
\begin{align}
    \label{eq:dimfree_results}
    \frac{1}{T}\sum_{t=1}^{T}\mathbb{E}\Big[\left\Vert\nabla f\left(\Bar{\mathbf{w}}^{t}\right)\right\Vert^{2}\Big] 
    \overset{(\text{a})}{\leq}\: \frac{1}{\Phi(\mathbf{A})T}\left(f(\Bar{\mathbf{w}}^{1}) - f(\Bar{\mathbf{w}}^{\star})\right) + \frac{\Theta_{2}}{\Phi(\mathbf{A})}\sum\limits_{n\in\mathcal{N}}\max\limits_{m\in\mathcal{M}}\left\{\left(\frac{a_{m,n}}{\sum\limits_{n'\in\mathcal{N}}a_{m,n'}}\right)^{2}\right\},
\end{align}
where inequality (a) is obtained by using $f(\bar{\mathbf{w}}^{T+1}) \geq f(\mathbf{w}^{\star})$.
This completes the proof of Theorem 1.
\end{proof}

\end{document}